\documentclass{article}

\usepackage{microtype}
\usepackage{graphicx}
\usepackage{subcaption}
\usepackage{booktabs} 
\usepackage{caption}
\usepackage{xr}

\usepackage{tikz}
\usepackage{amsfonts}%
\usepackage{amssymb}%
\usepackage{amsthm}
\usepackage{amsmath}
\usepackage{mathtools}
\usepackage{algorithm}
\usepackage[noend]{algorithmic}

\DeclareMathOperator{\dis}{dis}
\DeclareMathOperator{\doo}{do}

\DeclareMathOperator{\pa}{pa}
\DeclareMathOperator{\de}{de}
\DeclareMathOperator{\si}{si}
\DeclareMathOperator{\ch}{ch}
\DeclareMathOperator{\an}{an}
\DeclareMathOperator{\pas}{pa^s}

\DeclareMathOperator{\mb}{mb}

\DeclareMathOperator{\pre}{pre}

\def\ci{\perp\!\!\!\perp}

\theoremstyle{plain}

\newtheorem{thm}{Theorem}
\newtheorem{lem}{Lemma}

\theoremstyle{definition}
\newtheorem{dfn}{Definition}

\theoremstyle{remark}

\usepackage{hyperref}

\usepackage[accepted]{icml2020}

\icmltitlerunning{Identification Methods With Arbitrary Interventional Distributions as Inputs}

\begin{document}

\twocolumn[
\icmltitle{Identification Methods With Arbitrary Interventional Distributions as Inputs}




\begin{icmlauthorlist}
\icmlauthor{Jaron J. R. Lee}{jh}
\icmlauthor{Ilya Shpitser}{jh}
\end{icmlauthorlist}

\icmlaffiliation{jh}{Department of Computer Science, Johns Hopkins University, Baltimore, Maryland, USA}

\icmlcorrespondingauthor{Jaron J. R. Lee}{jaron.lee@jhu.edu}

\icmlkeywords{Machine Learning, Causal Inference}

\vskip 0.3in
]



\printAffiliationsAndNotice{}  

\begin{abstract}    
Causal inference quantifies cause-effect relationships by estimating counterfactual parameters from data.  This entails using \emph{identification theory} to establish a link between counterfactual parameters of interest and distributions from which data is available. A line of work characterized non-parametric identification for a wide variety of causal parameters in terms of the \emph{observed data distribution}. More recently, identification results have been extended to settings where experimental data from interventional distributions is also available. In this paper, we use Single World Intervention Graphs and a nested factorization of models associated with mixed graphs to give a very simple view of existing identification theory for experimental data.  We use this view to yield general identification algorithms for settings where the input distributions consist of an arbitrary set of observational and experimental distributions, including marginal and conditional distributions.  We show that for problems where inputs are interventional marginal distributions of a certain type (ancestral marginals), our algorithm is complete. 
\end{abstract}

\section{Introduction}

Causal inference quantifies cause-effect relationships using parameters associated with \emph{counterfactual responses} to an \emph{intervention operation}, where variables are set to values, possibly contrary to fact.  This operation is denoted by do$(.)$ in \cite{pearl09causality}.
In statistics and public health, counterfactual responses, or potential outcomes are denoted as $Y(\mathbf{a})$, which reads ``the variable $Y$ had the set of variables $\mathbf{A}$ been set to values $\mathbf{a}$.''
Cause-effect relationships are quantified by low dimensional parameters of counterfactual distributions.  For example, the \emph{average causal effect (ACE)} $\mathbb{E}[Y(\mathbf{a})] - \mathbb{E}[Y(\mathbf{a}')]$ quantifies the impact of treatment variables $\mathbf{A}$ on the outcome $Y$ by comparing means in a hypothetical randomized controlled trial where the treatments in one arm are set to $\mathbf{a}$, and in another arm to $\mathbf{a}'$.

Counterfactual outcomes $Y(\mathbf{a})$ are linked to factual outcomes $Y$ using the \emph{consistency property} which states that for any unit in the data where $\mathbf{A}$ is observed to equal $\mathbf{a}$, $Y(\mathbf{a}) = Y$.  However, values of $Y(\mathbf{a}')$ for such units are unobserved if $\mathbf{a}' \neq \mathbf{a}$, leading to the \emph{fundamental problem of causal inference}.  This problem is addressed by causal models, which use assumptions on the joint distribution of factual and counterfactual random variables to express desired causal parameters as functionals of the observed data distribution.

As a simple example (known in the literature as the conditional ignorability model), if observed variables include the treatment $A$, outcome $Y$, and a set of baseline covariates $\mathbf{C}$, and these covariates suffice to adjust for confounding (meaning that the conditional ignorability assumption $Y(a) \ci A \mid \mathbf{C}$ holds), then under the positivity condition $p(a \mid \mathbf{C}) > 0$ for all $a$, the ACE is identified by the \emph{adjustment formula}:
\begin{align*}
\mathbb{E}[Y(a)] - \mathbb{E}[Y(a')] = \mathbb{E}[ \mathbb{E}[Y | a, \mathbf{C}] - \mathbb{E}[Y | a', \mathbf{C}]].
\end{align*}

A complete theory has been developed that uses assumptions in a causal model to check which interventional distributions are identified, and express all identifiable interventional distributions as functionals of the observed data \cite{tian02on,shpitser06id, huang06ident}.  If the causal model yields the observed data distribution that admits a factorization with respect to a graph, then identified interventional distributions are always equal to \emph{modified factorizations} of an appropriate graphical model representing the observed data distribution \cite{lauritzen02chain,richardson17nested}.

A natural generalization considered the problem of identification from \emph{surrogate experiments} where a target causal parameter is expressed in terms of a set of distributions arising from performing experiments on a particular population (including possibly the ``null experiment,'' which recovers the observed data distribution).  A line of work gave increasingly general identification algorithms for this problem \cite{bareinboim12z, leeGeneralIdentifiabilityArbitrary2019}.

In this paper, we show that Single World Intervention Graphs (SWIGs) \cite{thomas13swig}, and the nested factorization of mixed graphs \cite{richardson17nested} yield a very simple view of the theory of identification from experimental data.  We use this view to give a series of general
algorithms for identification in terms of \emph{arbitrary sets} of observational or interventional distributions.  In addition, we show that for a particular class of inputs, our algorithm is complete.

\begin{figure*}
    \captionsetup{format=hang}
    \centering
    \begin{subfigure}[b]{0.12\textwidth}
        \begin{tikzpicture}[
            > = stealth, 
            auto,
            semithick 
            ]

            \tikzstyle{state}=[
            draw = none,
            fill = white,
            minimum size = .8mm
            ]

            \node[state] (X1) {$X_1$};
            \node[state] (X2) [right of=X1] {$X_2$};
            \node[state] (U) [below of=X2] {$U$};
            \node[state] (W) [below of=X1] {$W$};
            \node[state] (Y) [below of=U] {$Y$};

            \path[<->, red] (X1) edge node {} (X2);
            \path[->, blue] (X1) edge node {} (W);
            \path[->, blue] (W) edge node {} (Y);
            \path[->, blue] (X2) edge node {} (U);
            \path[->, blue] (U) edge node {} (Y);
            \path[<->, red, bend right=30] (X1) edge node {} (W);
            \path[<->, red, bend right=30] (U) edge node {} (Y);
            \path[<->, red, bend left=30] (X2) edge node {} (Y);

        \end{tikzpicture}
        \caption{$\mathcal{G}$}
        \label{fig:latentu1}
    \end{subfigure}
    \begin{subfigure}[b]{0.12\textwidth}
        \begin{tikzpicture}[
            > = stealth, 
            auto,
            semithick 
            ]

            \tikzstyle{state}=[
            draw = none,
            fill = white,
            minimum size = 2mm
            ]
            \tikzstyle{fixed}=[
            draw=black,
            rectangle, 
            thick,
            fill = white,
            minimum size = 2mm
            ]

            \node[fixed] (X1) {$x_1$};
            \node[state] (X2) [right of=X1] {$X_2$};
            \node[state] (U) [below of=X2] {$U$};
            \node[state] (W) [below of=X1] {$W(x_1)$};
            \node[state] (Y) [below of=U] {$Y(x_1)$};

            \path[->, blue] (X1) edge node {} (W);
            \path[->, blue] (W) edge node {} (Y);
            \path[->, blue] (X2) edge node {} (U);
            \path[->, blue] (U) edge node {} (Y);
            \path[<->, red, bend right=30] (U) edge node {} (Y);
            \path[<->, red, bend left=30] (X2) edge node {} (Y);

        \end{tikzpicture}
        \caption{${\cal G}_1$
        }
        \label{fig:latentu2}
    \end{subfigure}
    \begin{subfigure}[b]{0.12\textwidth}
        \begin{tikzpicture}[
            > = stealth, 
            auto,
            semithick 
            ]

            \tikzstyle{state}=[
            draw = none,
            fill = white,
            minimum size = 2mm
            ]
            \tikzstyle{fixed}=[
            draw=black,
            rectangle, 
            thick,
            fill = white,
            minimum size = 2mm
            ]

            \node[state] (X1) {$X_1$};
            \node[fixed] (X2) [right of=X1] {$x_2$};
            \node[state] (U) [below of=X2] {$U(x_2)$};
            \node[state] (W) [below of=X1] {$W$};
            \node[state] (Y) [below of=U] {$Y(x_2)$};

            \path[->, blue] (X1) edge node {} (W);
            \path[->, blue] (W) edge node {} (Y);
            \path[->, blue] (X2) edge node {} (U);
            \path[->, blue] (U) edge node {} (Y);
            \path[<->, red, bend right=30] (X1) edge node {} (W);
            \path[<->, red, bend right=30] (U) edge node {} (Y);

        \end{tikzpicture}
        \caption{
        ${\cal G}_2$
        }
        \label{fig:latentu3}
    \end{subfigure}
    \begin{subfigure}[b]{0.12\textwidth}
        \begin{tikzpicture}[
            > = stealth, 
            auto,
            semithick 
            ]

            \tikzstyle{state}=[
            draw = none,
            fill = white,
            minimum size = 2mm
            ]
            \tikzstyle{fixed}=[
            draw=black,
            rectangle, 
            thick,
            fill = white,
            minimum size = 2mm
            ]

            \node[fixed] (X1) {$x_1$};
            \node[state] (W) [below of=X1] {$W(x_1)$};

            \path[->, blue] (X1) edge node {} (W);

        \end{tikzpicture}
        \caption{
        ${\cal G}_3$
        }
        \label{fig:latentu4}
    \end{subfigure}
%
%
%
%
    \begin{subfigure}[b]{0.12\textwidth}
        \begin{tikzpicture}[
            > = stealth, 
            auto,
            semithick 
            ]

            \tikzstyle{state}=[
            draw = none,
            fill = white,
            minimum size = 2mm
            ]
            \tikzstyle{fixed}=[
            draw=black,
            rectangle, 
            thick,
            fill = white,
            minimum size = 2mm
            ]

            \node[fixed] (X2) [right of=X1] {$x_2$};
            \node[state] (W) [below of=X1] {$W$};
            \node[state] (Y) [below of=U] {$Y(x_2)$};

            \path[->, blue] (W) edge node {} (Y);
            \path[->, blue] (X2) edge node {} (Y);

        \end{tikzpicture}
        \caption{
        ${\cal G}_4$
        }
        \label{fig:latentu6}
    \end{subfigure}\hspace{9mm}%
    \begin{subfigure}[b]{0.10\textwidth}
        \begin{tikzpicture}[
            > = stealth, 
            auto,
            semithick 
            ]

            \tikzstyle{state}=[
            draw = none,
            fill = white,
            minimum size = 1mm
            ]
            \tikzstyle{fixed}=[
            draw=black,
            rectangle, 
            thick,
            fill = white,
            minimum size = 1mm
            ]

            \node[fixed] (X1) {$x_1$};
            \node[fixed] (X2) [right of=X1, xshift=0.2cm] {$x_2$};
            \node[state] (U) [below of=X2] {$U(x_2)$};
            \node[state] (W) [below of=X1] {$W({x_1})$};
            \node[state] (Y) [below of=U] {$Y({x_1,x_2})$};

            \path[->, blue] (X1) edge node {} (W);
            \path[->, blue] (W) edge node {} (Y);
            \path[->, blue] (X2) edge node {} (U);
            \path[->, blue] (U) edge node {} (Y);
             \path[<->, red, bend right=30] (U) edge node {} (Y);

        \end{tikzpicture}
        \caption{
        ${\cal G}_5$
        }
        \label{fig:latentu7}
    \end{subfigure}
    \label{fig:latentu}
    \caption{A graph $\mathcal{G}$ and corresponding interventional distributions for the example given in Section \ref{sec:motivating}.
    }
\end{figure*}
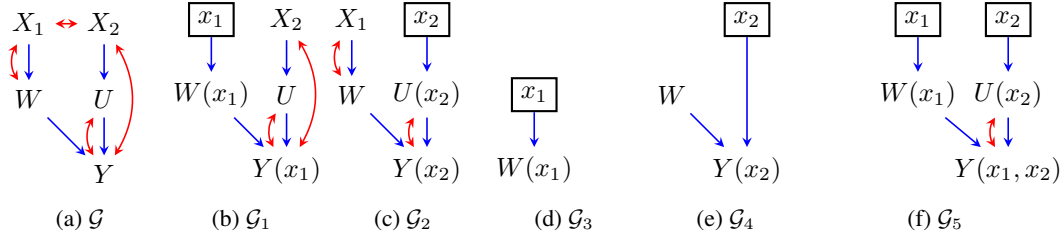

\subsection{Motivating Example}\label{sec:motivating} 

Consider a causal model represented by a graph shown in Fig.~\ref{fig:latentu1}, where directed edges denote causation, and bidirected edges denote the presence of a hidden common cause
\footnote{This example is inspired by Fig.~1 in \citet{leeGeneralIdentifiabilityArbitrary2019}.}.
Here $Y$ is a health outcome, namely the presence of cardiovascular disease, $X_1$ denotes whether a hip replacement was performed, and $X_2$ denotes whether an atrial valve replacement was performed.  Furthermore, let $W$ denote the ability to walk (influenced by whether hip replacement was performed), and $U$ denote an aspect of heart health (such as valve regurgitation). Hip problems and heart problems do not have any direct causal connection, but are certainly confounded by a patient's general health. Plausibly, a hip replacement which causes a patient to be unable to walk would certainly impact their overall health, and could contribute to the development of cardiovascular disease.  Additionally, the measure $U$ and heart disease $Y$ is confounded by the doctor's latent knowledge of the patient's health.

We want to learn how hip surgery and valve replacement surgeries affect cardiovascular disease by considering the distribution $p(Y(x_1, x_2))$.  Given the type of unobserved confounding present in the problem, existing results in \citet{shpitser06id} imply that this distribution is not identified from the observed data distribution $p(X_1, X_2, W, U, Y)$.

However, suppose that we have access to a data set where patients (from the population we wished to consider) elected to be randomized to hip replacement versus a non-invasive alternative treatment, and to another data set whre patients elected to be randomized to valve replacement versus a non-invasive alternative treatment.  Data from these RCTs is represented by interventional distributions
$p(Y(x_1), X_2(x_1), W(x_1), U(x_1), Y(x_1))$, and $p(Y(x_2), X_1(x_2), W(x_2), U(x_2), Y(x_2))$, respectively.  Graphs representing these two RCTs, called \emph{mutilated graphs} \cite{pearl09causality}, are shown in Figs.~\ref{fig:latentu2} and ~\ref{fig:latentu3}.  The gID algorithm, described in \citet{leeGeneralIdentifiabilityArbitrary2019} is able to identify the target distribution $p(X_1, X_2, W, U, Y)$ from the two input interventional distributions above.

While gID was proven sound and complete, it has the limitation that requires that every input distribution contains every observed variable: either as an outcome, or an intervened-on treatment.  This limits its utility for certain types of causal inference problems, as we now illustrate.
Suppose that the available RCTs were performed by separate research groups with differing data collection policies. For example, the RCT studying hip surgery was scoped only for its impact on walking ability, yielding the distribution $p(W(x_1))$, represented by Fig. ~\ref{fig:latentu4}
In practice, we should not expect all studies used for analysis to contain all variables relevant in the problem.  Marginal distributions are not valid inputs for gID, and require a more general algorithm.

In this paper, we consider extensions to the gID algorithm that are able to identify interventional distributions given increasingly arbitrary interventional distributions as inputs.  We build up to the most general algorithm by considering how gID generalizes in a number of special cases.

The paper is organized as follows.  We introduce necessary preliminaries in Section ~\ref{sec:prelim}, reformulate the existing identification algorithm for experimental distribution inputs \cite{leeGeneralIdentifiabilityArbitrary2019}, as well as our generalizations in Section ~\ref{sec:main}, and describe completeness results in Section ~\ref{sec:complete}.  Section ~\ref{sec:conclusion} contains our conclusions.  We defer proofs of all results to the Appendix.

\section{Graphs and Graphical Models}
\label{sec:prelim}
Let capital letters $X$ denote random variables, and let lower case letters $x$ values. Sets of random variables are denoted $\mathbf{V}$, and sets of values $\mathbf{v}$.
For a subset $\mathbf{A} \subseteq \mathbf{V}$, $\mathbf{v}_{\mathbf{A}}$ denotes the subset of values in $\mathbf{v}$ of variables in $\mathbf{A}$.
Domains of $X$ and $\mathbf{X}$ are denoted by ${\mathfrak X}_X$ and ${\mathfrak X}_{\mathbf{X}}$, respectively.

We use standard genealogic relations on graphs: parents, children, descendants, siblings and ancestors of $X$ in a graph ${\cal G}$ are denoted by
$\pa_{\cal G}(X), \ch_{\cal G}(X), \de_{\cal G}(X), \si_{\cal G}(X), \an_{\cal G}(X)$, respectively \cite{lauritzen96graphical}.  These relations are defined disjunctively for sets, e.g.
$\pa_{\cal G}(\mathbf{X}) \equiv \bigcup_{X \in \mathbf{X}} \pa_{\cal G}(X)$.
We will also define the set of \emph{strict parents} as follows: $\pas_{\cal G}(\mathbf{X}) = \pa_{\cal G} (\mathbf{X}) \setminus \mathbf{X}$.
Given any vertex $V$ in an ADMG ${\cal G}$, define the \emph{ordered Markov blanket} of $V$ as $\mb_{\cal G}(V) \equiv (\dis_{\cal G}(V) \cup \pa_{\cal G}(\dis_{\cal G}(V)) ) \setminus \{V\}$.
Given a graph ${\cal G}$ with vertex set $\mathbf{V}$, and $\mathbf{S} \subseteq \mathbf{V}$, define the \emph{induced subgraph} $\mathcal{G}_{\mathbf{S}}$ to be a graph containing the vertex set $\mathbf{S}$ and all edges in ${\cal G}$ among elements in $\mathbf{S}$.

We will consider directed acyclic graphs (DAGs), which are graphs with directed edges and no directed cycles, and acyclic directed mixed graphs (ADMGs), which are graphs with directed and bidirected edges and no directed cycles.
A bidirected connected component in an ADMG is called a \emph{district} (also known as a c-component).  A set of districts of an ADMG ${\cal G}$ with vertex set $\mathbf{V}$, which we will denote by ${\cal D}({\cal G})$, partitions $\mathbf{V}$.
The district of $V \in \mathbf{V}$ in ${\cal G}(\mathbf{V})$ is denoted by $\dis_{{\cal G}(\mathbf{V})}(V)$.
By convention, for any $X$, $\an_{\cal G}(X) \cap \de_{\cal G}(X) \cap \dis_{\cal G}(X) = \{ X \}$.

A statistical model of DAG ${\cal G}$ with vertex set $\mathbf{V}$ is the set of all distributions $p(\mathbf{V})$ that Markov-factorize according to ${\cal G}$, as follows:
$p(\mathbf{V}) = \prod_{V \in \mathbf{V}} p(V \mid \pa_{\cal G}(V))$.

Causal models are sets of distributions on counterfactual random variables. For some $Y \in \mathbf{V}$, $\mathbf{A} \subseteq \mathbf{V} \setminus \{Y \}$, a counterfactual random variable $Y(\mathbf{a})$ reads ``value of $Y$ had $\mathbf{A}$ been set, possibly contrary to fact, to $\mathbf{a}$.''
For convenience we will denote distributions over multiple counterfactuals $p(Y_1(\mathbf{a}), \ldots, Y_k(\mathbf{a})))$ as $p(\{Y_1, \ldots, Y_k\}(\mathbf{a}))$, or $p(\mathbf{Y}(\mathbf{a}))$ if $\mathbf{Y} \equiv \{ Y_1, \ldots, Y_k \}$.  The same distribution had been denoted by $p(\mathbf{Y} | \text{do}(\mathbf{a}))$ in \cite{pearl09causality}.

Causal models of a DAG $\mathcal{G}(\mathbf{V})$ 
are defined on counterfactual variables $V(\mathbf{a}_V)$, for all $\mathbf{a}_V \in {\mathfrak X}_{\pa_{\mathcal{G}}(V)}$.
In this paper, we use Pearl's functional model for a DAG $\mathcal{G}(\mathbf{V})$, which is defined by the restriction that the sets of variables
$\{V(\mathbf{a_V}) \mid \mathbf{a}_V \in \mathfrak{X}_{\pa_{\mathcal{G}} (V)}\}$ for every $V \in \mathbf{V}$ are mutually independent.
Under this model, for every $\mathbf{A}$, the distribution $p(\mathbf{V}(\mathbf{a}))$ is identified by a modified DAG factorization known as the \emph{extended g-formula}:
$\prod_{V \in \mathbf{V}} p(V | \mathbf{a}_{\pa_{\cal G}(V) \cap \mathbf{A}}, \pa_{\cal G}(V) )$.

Conditional independences in $p(\mathbf{V}(\mathbf{a}))$ implied by a causal DAG model may be read off from a special DAG called a Single World Intervention Graph (SWIG).
Given a set $\mathbf{A} \subseteq \mathbf{V}$ of variables and an assignment $\mathbf{a}$ to those variables, a SWIG ${\cal G}(\mathbf{V}(\mathbf{a}))$ is constructed from ${\cal G}(\mathbf{V})$ by splitting all vertices in $\mathbf{A}$ into a random half and a fixed half, with the random half inheriting all edges with an incoming arrowhead and the fixed half inheriting all outgoing directed edges. Then, all random vertices $V_i$ are re-labelled as $V_i(\mathbf{a})$ or equivalently 
as $V_i(\mathbf{a}_i)$,
where $\mathbf{a}_i$ consists of the values of fixed fixed vertices that are ancestors of $V_i$ in the split graph; the latter labelling is referred to as the \emph{minimal labelling} of the SWIG.
Under standard causal models of a DAG, the interventional distribution $p(\mathbf{V}(\mathbf{a}))$ factorizes as follows with respect to the SWIG ${\cal G}(\mathbf{V}(\mathbf{a}))$:
\begin{align*}
    \prod_{V(\mathbf{a}) \in \mathbf{V}(\mathbf{a})} p(V(\mathbf{a}) | \{ W(\mathbf{a}) : W \in \pa_{{\cal G}(\mathbf{V}(\mathbf{a}))}(V(\mathbf{a})) \setminus \mathbf{a} \}),
\end{align*}
where each 
$p(V(\mathbf{a}) | \{ W(\mathbf{a}) : W \in \pa_{{\cal G}(\mathbf{V}(\mathbf{a}))}(V(\mathbf{a})) \setminus \mathbf{a} \})$ is only a function of $\mathbf{a}$ that are parents of $V(\mathbf{a})$ (this qualification is substantive and potentially defines restrictions).  This factorization allows us to use standard d-separation relations on the SWIG ${\cal G}(\mathbf{V}(\mathbf{a}))$ (that potentially allow one of the endpoints to be a fixed vertex, and treat all other fixed vertices as conditioned on) to discover conditional independence or exclusion restrictions on $p(\mathbf{V}(\mathbf{a}))$. See \citet{thomas13swig, malinsky19po} for more details.

Most causal models in practice contain hidden variables, which significantly complicates identification theory.  An interventional distribution $p(\mathbf{Y}(\mathbf{a}))$ may not be identified at all from the observed marginal distribution in hidden variable models.  However, if $p(\mathbf{Y}(\mathbf{a}))$ is identified, it is equal to a modified factorization of a graphical model associated with a certain mixed graph, just as identified $p(\mathbf{Y}(\mathbf{a}))$ are equal to a modified DAG factorization if the causal model is fully observed.

A hidden variable causal model of a DAG is represented by a DAG ${\cal G}$ with vertices $\mathbf{V} \cup \mathbf{H}$, with $\mathbf{V}$ representing observed variables, and $\mathbf{H}$ representing hidden variables.  Given a hidden variable DAG ${\cal G}(\mathbf{V} \cup \mathbf{H})$, the mixed graph we will be interested in is called a \emph{latent projection} of ${\cal G}(\mathbf{V} \cup \mathbf{H})$ on $\mathbf{V}$, and will be denoted by ${\cal G}(\mathbf{V})$ (by analogy with marginal distribution notation in probability theory).  ${\cal G}(\mathbf{V})$ is an ADMG with vertices $\mathbf{V}$, a directed edge between any $V_i,V_j$ linked by a directed path in ${\cal G}(\mathbf{V} \cup \mathbf{H})$ where all intermediate vertices are in $\mathbf{H}$, and a bidirected edge between any $V_i,V_j$ linked by a marginally d-connected path in ${\cal G}(\mathbf{V} \cup \mathbf{H})$ where all intermediate vertices are in $\mathbf{H}$, the first edge is directed into $V_i$, and the last edge is directed into $V_j$.

The latent projection operation generalizes in the natural way to SWIGs.
Just as a latent projection ${\cal G}({\bf V})$ of a hidden variable DAG ${\cal G}({\bf V} \cup {\bf H})$ represents the structure of a marginal distribution $p({\bf V}))$ obtained from $p({\bf V} \cup {\bf H})$, so does a latent projection ${\cal G}({\bf V})$ of a SWIG  ${\cal G}(\{ {\bf V} \cup {\bf H} \}({\bf a}))$ represents the structure of a marginal counterfactual distribution $p({\bf V}({\bf a}))$ obtained from
$p(\{ {\bf V} \cup {\bf H} \}({\bf a}))$.

A marginal SWIG ${\cal G}({\bf V}({\bf a}))$ may be constructed from the SWIG ${\cal G}(\{ \mathbf{V} \cup \mathbf{H} \}(\mathbf{a}))$ (itself constructed from a latent variable DAG ${\cal G}(\mathbf{V} \cup \mathbf{H})$ by splitting vertices in $\mathbf{A}$) by ``projecting out'' variables $\mathbf{H}(\mathbf{a})$ using the standard latent projection construction. Note that the operations of splitting vertices that yield SWIGs, and projecting out vertices corresponding to hidden variables commute provided all split vertices correspond to observed variables \cite{malinsky19po}.

A hidden variable DAG ${\cal G}(\mathbf{V} \cup \mathbf{H})$ may be used to define a factorization on marginal distributions $p(\mathbf{V})$ in terms of the DAG as: $p(\mathbf{V}) = \sum_{\mathbf{H}} \prod_{V \in \mathbf{V} \cup \mathbf{H}} p(V | \pa_{\cal G}(V))$.  However, such a factorization is difficult to work with in causal inference applications, since the corresponding likelihood is difficult to specify correctly, and leads to a model with singularities \cite{drton09likelihood}.

A principled alternative is to define a factorization of a marginal distribution $p(\mathbf{V})$ directly on the latent projection ADMG ${\cal G}(\mathbf{V})$.  Such a \emph{nested Markov factorization}, described in \citet{richardson17nested} completely avoids modeling hidden variables, and leads to a regular likelihood in special cases \cite{evans18smooth,shpitser18sem}, while capturing all equality constraints a hidden variable DAG factorization imposes on the observed margin $p(\mathbf{V})$ \cite{evansMarginsDiscreteBayesian2018}.  In addition, $p(\mathbf{Y}(\mathbf{a}))$ identified in a hidden variable causal model represented by ${\cal G}(\mathbf{V} \cup \mathbf{H})$ is always equal to a modified version of a nested factorization \cite{richardson17nested} associated with ${\cal G}(\mathbf{V})$, which we briefly describe.  

\subsection{The Nested Markov Factorization}

The nested factorization of $p(\mathbf{V})$ with respect to an ADMG ${\cal G}(\mathbf{V})$ links \emph{Markov kernels} derived from $p(\mathbf{V})$ to \emph{conditional graphs} derived from ${\cal G}(\mathbf{V})$ via a graphical and probabilistic \emph{fixing operator}.

A conditional ADMG (CADMG) is a graph ${\cal G}(\mathbf{V},\mathbf{W})$ with random vertices $\mathbf{V}$, and fixed vertices $\mathbf{W}$, directed and bidirected edges, no directed cycles, and no edges with arrowheads into any element of $\mathbf{W}$.
All genealogic relations transfer from ADMGs to CADMGs without change, except districts in a CADMG are defined only on the set $\mathbf{V}$.
A CADMG without bidirected edges is called a conditional DAG (CDAG).

A kernel $q_{\mathbf{V}}(\mathbf{V} | \mathbf{W})$ is a mapping from ${\mathfrak X}_{\mathbf{W}}$ to normalized densities over $\mathbf{V}$.  A conditional distribution is a kernel, although some kernels are not conditional distributions -- for example, $q_{Y}(Y | a) = \sum_{C} p(Y | a, C) p(C)$ is a kernel arising under conditional ignorability
that is not equal to $p(Y | a)$ unless $A$ is marginally independent of $C$.  For any $\mathbf{A} \subseteq \mathbf{V}$, we define the following shorthand notation:
$q_{\mathbf{V}}(\mathbf{A} | \mathbf{W}) \equiv \sum_{{\mathbf V} \setminus \mathbf{A}} q_{{\mathbf V}}(\mathbf{V} | \mathbf{W})$, $q_{\mathbf{V}} (\mathbf{V} \setminus \mathbf{A} | \mathbf{A},\mathbf{W}) \equiv q_{{\mathbf V}}(\mathbf{V} | \mathbf{W}) / q_{\mathbf{V}}(\mathbf{A} | \mathbf{W})$.

$V \in \mathbf{V}$ is said to be fixable in ${\cal G}(\mathbf{V},\mathbf{W})$ if $\de_{\cal G}(V) \cap \dis_{\cal G}(V) = \emptyset$.  We define a fixing operator $\phi_V({\cal G})$ for graphs, and a fixing operator $\phi_V(q; {\cal G})$ for kernels.

Given a CADMG ${\cal G}(\mathbf{V},\mathbf{W})$, with a fixable $V \in \mathbf{V}$, $\phi_V({\cal G}(\mathbf{V},\mathbf{W}))$ yields a new CADMG ${\cal G}(\mathbf{V} \setminus \{ V \}, \mathbf{W} \cup \{ V \})$ obtained from ${\cal G}(\mathbf{V}, \mathbf{W})$ by moving $V$ from $\mathbf{V}$ to $\mathbf{W}$, and removing all edges with arrowheads into $V$.  Given a kernel $q_{\mathbf{V}}(\mathbf{V} | \mathbf{W})$, and a CADMG ${\cal G}(\mathbf{V}, \mathbf{W})$, $\phi_V(q_{\mathbf{V}}(\mathbf{V} | \mathbf{W}), {\cal G}(\mathbf{V}, \mathbf{W}))$ yields a new kernel:
\[
    q_{\mathbf{V} \setminus \{ V \}}(\mathbf{V} \setminus \{ V \} | \mathbf{W} \cup \{ V \}) \equiv
    \frac{
        q_{\mathbf{V}}(\mathbf{V} | \mathbf{W})
        }{
        q_{\mathbf{V}}(V | \mb_{\cal G}(V), \mathbf{W})
    }.
\]

A sequence $\langle V_1, \ldots, V_k \rangle$ is said to be \emph{valid} in ${\cal G}(\mathbf{V},\mathbf{W})$ if $V_1$ fixable in ${\cal G}(\mathbf{V}, \mathbf{W})$, $V_2$ is fixable in $\phi_{V_1}({\cal G}(\mathbf{V}, \mathbf{W}))$, and so on.  If any two sequences $\sigma_1, \sigma_2$ for the same set $\mathbf{S} \subseteq \mathbf{V}$ are fixable in ${\cal G}$, they lead to the same CADMG.  As a result, we extend the graph fixing operator to a set $\mathbf{S}$: $\phi_{\mathbf{S}}({\cal G})$. This operator is defined as applying the vertex fixing operation in any valid sequence on elements in $\mathbf{S}$.

Given a sequence $\sigma_{\mathbf{S}}$, define $\eta(\sigma_{\mathbf{S}})$ to be the first element in $\sigma_{\mathbf{S}}$, and $\tau(\sigma_{\mathbf{S}})$ to be the subsequence of $\sigma_{\mathbf{S}}$ containing all but the first element.

We extend the kernel fixing operator to \emph{sequences}: given a sequence $\sigma_{\mathbf{S}}$ on elements in $\mathbf{S}$ valid in ${\cal G}(\mathbf{V}, \mathbf{W})$,
$\phi_{\sigma_{\mathbf{S}}}(q_{\mathbf{V}}(\mathbf{V} | \mathbf{W}), {\cal G}(\mathbf{V}, \mathbf{W}))$ is defined to be equal to $q_{\mathbf{V}}(\mathbf{V} | \mathbf{W})$ if $\sigma_{\mathbf{S}}$ is the empty sequence, and
$\phi_{\tau(\sigma_{\mathbf{S}})}( \phi_{\eta(\sigma_{\mathbf{S}})}(q_{\mathbf{V}}(\mathbf{V} | \mathbf{W}); {\cal G}(\mathbf{V}, \mathbf{W})), \phi_{\eta(\sigma_{\mathbf{S}})}({\cal G}(\mathbf{V}, \mathbf{W})))$ otherwise.

Given a CADMG ${\cal G}(\mathbf{V}, \mathbf{W})$, a set $\mathbf{R} \subseteq \mathbf{V}$ is called \emph{reachable} if there exists a sequence for $\mathbf{V} \setminus \mathbf{R}$ valid in ${\cal G}(\mathbf{V},\mathbf{W})$.  A set $\mathbf{R}$ reachable in ${\cal G}(\mathbf{V},\mathbf{W})$ is \emph{intrinsic} in ${\cal G}(\mathbf{V},\mathbf{W})$ if $\phi_{\mathbf{V} \setminus \mathbf{R}}({\cal G})$ contains a single district, $\mathbf{R}$ itself.  The set of intrinsic sets in a CADMG ${\cal G}$ is denoted by ${\cal I}({\cal G})$.

A distribution $p(\mathbf{V})$ is said to obey the \emph{nested Markov factorization} with respect to the ADMG ${\cal G}(\mathbf{V})$ if there exists a set of kernels of the form $\{ q_{\mathbf{S}}(\mathbf{S} | \pas_{\cal G}(\mathbf{S}) ) : \mathbf{S} \in {\cal I}({\cal G}) \} \}$ such that for every valid sequence $\sigma_{\mathbf{R}}$ for a reachable set $\mathbf{R}$ in ${\cal G}$, we have:
\[
    \phi_{\sigma_{\mathbf{R}}}(p(\mathbf{V}); {\cal G}(\mathbf{V})) = \prod_{\mathbf{D} \in {\cal D}(\phi_{\mathbf{R}}({\cal G}(\mathbf{V})))} q_{\mathbf{D}}(\mathbf{D} | \pas_{\cal G}(\mathbf{D}) ).
\]
If a distribution obeys this factorization, then for any reachable $\mathbf{R}$, any two valid sequences on $\mathbf{R}$ applied to $p(\mathbf{V})$ yield the same kernel $q_{\mathbf{R}}(\mathbf{R} | \mathbf{V} \setminus \mathbf{R})$.  Hence, kernel fixing may be defined on sets, just as graph fixing.  In this case, for every
$\mathbf{D} \in {\cal I}({\cal G})$, $q_{\mathbf{D}}(\mathbf{D} | \pas_{\cal G}(\mathbf{D}) ) \equiv \phi_{\mathbf{V} \setminus \mathbf{D}}(p(\mathbf{V}); {\cal G}(\mathbf{V}))$.

One of the consequences of the nested factorization is the so called \emph{district factorization} or \emph{c-component factorization}:
\begin{align*}
    p(\mathbf{V})
&= \prod_{\mathbf{D} \in {\cal D}({\cal G}(\mathbf{V}))} q_{\mathbf{D}}(\mathbf{D} | \pas_{\cal G}(\mathbf{D})) \\
&
= \prod_{\mathbf{D} \in {\cal D}({\cal G}(\mathbf{V}))} \left(
\prod_{D \in \mathbf{D}} p(D | \pre_{\prec}(D))
\right),
\end{align*}
where $\pre_{\prec}(D)$ is the set of predecessors of $D$ according to a topological total ordering $\prec$.  Note that each factor $\prod_{D \in \mathbf{D}} p(D \mid \pre_{\prec}(D))$ is only a function of $\mathbf{D} \cup \pa_{\cal G}(\mathbf{D})$ under the nested factorization.

If $p(\mathbf{V} \cup \mathbf{H})$ Markov factorizes relative to a DAG ${\cal G}(\mathbf{V} \cup \mathbf{H})$, then the marginal distribution $p(\mathbf{V})$ nested Markov factorizes relative to the latent projection ADMG ${\cal G}(\mathbf{V})$.
A global Markov property has been defined for models obeying this factorization \cite{richardson17nested}, and it is known to logically imply all equality constraints imposed on a marginal distribution by a hidden variable DAG.

It is known that in a hidden variable causal model, not every interventional distribution $p(\mathbf{Y}(\mathbf{a}))$ is identified.  However,
\emph{every} $p(\mathbf{Y}(\mathbf{a}))$ identified from $p(\mathbf{V})$ can be expressed as a modified nested factorization as follows:
\begin{align*}
    p(\mathbf{Y}(\mathbf{a}))
&=\!\!
\sum_{\mathbf{Y}^* \setminus \mathbf{Y}} \prod_{\mathbf{D} \in {\cal D}({\cal G}({\mathbf{Y}^*}(\mathbf{a})))} p(\mathbf{D} | \doo(\pas_{\cal G}(\mathbf{D}) )) \vert_{\mathbf{A} = \mathbf{a}}\\
&=\!\!
\sum_{\mathbf{Y}^* \setminus \mathbf{Y}} \prod_{\mathbf{D} \in {\cal D}({\cal G}({\mathbf{Y}^*}(\mathbf{a})))} \phi_{\mathbf{V} \setminus \mathbf{D}}(p(\mathbf{V}); {\cal G}(\mathbf{V})) \vert_{\mathbf{A} = \mathbf{a}},
\end{align*}
where $\mathbf{Y}^* \equiv \an_{{\cal G}(\mathbf{V}(\mathbf{a}))}(\mathbf{Y}) \setminus \mathbf{a}$, and ${\cal G}({\bf Y}^*({\bf a}))$ is the latent projection of the SWIG ${\cal G}({\bf V}({\bf a}))$ onto ${\bf Y}^*({\bf a})$.  This modified factorization yields a particularly simple view of the ID algorithm \cite{tian02on,shpitser06id}, one that we extend to identification algorithms that treat interventional distributions as known inputs.

\section{Algorithms for Identification from Interventional Distribution Inputs}
\label{sec:main}

A SWIG ${\cal G}(\mathbf{V}(\mathbf{a}))$ obtained from a DAG ${\cal G}(\mathbf{V})$ may be viewed as a conditional DAG with random vertices $\mathbf{V}$ and fixed vertices $\mathbf{a}$.  Similarly, a marginal SWIG ${\cal G}(\mathbf{V}(\mathbf{a}))$ is a conditional ADMG with random vertices $\mathbf{V}$ and fixed vertices $\mathbf{a}$.
Definitions of fixability, as well as reachable and intrinsic sets carry over to marginal SWIGs without change.
In fact, by a simple extention of Lemma 56 of \cite{richardson17nested}, we can show that if $p(\{ \mathbf{V}\cup \mathbf{H}\}(\mathbf{a}))$ factorizes with respect to a SWIG ${\cal G}(\{\mathbf{V}\cup \mathbf{H}\}(\mathbf{a}))$, then
$p(\mathbf{V}(\mathbf{a}))$ admits the following nested SWIG factorization with respect to the latent projection SWIG ${\cal G}(\mathbf{V}(\mathbf{a}))$.  For every set $\mathbf{R} \subseteq \mathbf{V}$ reachable in ${\cal G}(\mathbf{V}(\mathbf{a}))$, the kernel $\phi_{\mathbf{V} \setminus \mathbf{R}}(p(\mathbf{V}(\mathbf{a})); {\cal G}(\mathbf{V}(\mathbf{a})))$ factorizes as
\begin{align*}
    \prod_{\mathbf{D} \in {\cal D}(\phi_{\mathbf{V} \setminus \mathbf{R}}({\cal G}(\mathbf{V}(\mathbf{a}))))} q_{\mathbf{D}}(\mathbf{D}(\mathbf{a}) | \pas_{{\cal G}(\mathbf{V}(\mathbf{a}))}(\mathbf{D}(\mathbf{a})) )
    \\
    = \prod_{\mathbf{D} \in {\cal D}(\phi_{\mathbf{V} \setminus \mathbf{R}}({\cal G}(\mathbf{V}(\mathbf{a}))))} \phi_{\mathbf{V} \setminus \mathbf{D}}(p(\mathbf{V}(\mathbf{a})); {\cal G}(\mathbf{V}(\mathbf{a}))),
\end{align*}
where each term $q_{\mathbf{D}}(\mathbf{D}(\mathbf{a}) | \pas_{{\cal G}(\mathbf{V}(\mathbf{a}))}(\mathbf{D}(\mathbf{a})) )$ is only a function of those elements of $\mathbf{a}$ that are in $\pa_{{\cal G}(\mathbf{V}(\mathbf{a}))}(\mathbf{D}(\mathbf{a}))$.  These terms correspond to the set of intrinsic sets in the SWIG ${\cal I}({\cal G}({\bf V}({\bf a})))$.

In other words, under standard causal models of a DAG with hidden variables ${\cal G}(\mathbf{H} \cup \mathbf{V})$, marginal SWIGs ${\cal G}(\mathbf{S}(\mathbf{a}))$ represent represent structure of a marginal counterfactual $p(\mathbf{S}(\mathbf{a}))$, for any $\mathbf{S} \subseteq \mathbf{V}$ using the SWIG version of the nested Markov factorization.

\subsection{The gID algorithm as a modified nested factorization with respect to a set of SWIGs}

Checking if 
$p(\mathbf{Y}(\mathbf{a}))$ is identified is equivalent to checking whether 
$p(\mathbf{Y}^*(\mathbf{a})) = \prod_{\mathbf{D} \in {\cal D}({\cal G}({\mathbf{Y}^*}(\mathbf{a})))} p(\mathbf{D}(\pas_{\cal G}(\mathbf{D}) )) \vert_{\mathbf{A} = \mathbf{a}}$ is identified,
where
$\mathbf{Y}^*  \equiv  \an_{{\cal G}(\mathbf{V}(\mathbf{a}))}(\mathbf{Y}) \setminus \mathbf{a}$. 

The ID algorithm described above simply checks whether each district $\mathbf{D}$ in the marginal SWIG
${\cal G}(\mathbf{Y}^*(\mathbf{a}))$ corresponds to an intrinsic set in ${\cal G}(\mathbf{V})$. If so, it obtains the corresponding distribution $p(\mathbf{D}(\pa_{\cal G}(\mathbf{D})))$ via the appropriate factor in the nested Markov factorization, which in turn is a functional of $p(\mathbf{V})$ obtained by the fixing operator $\phi(.)$.  If not, the distribution $p(\mathbf{Y}(\mathbf{a}))$ turns out not to be identified from $p(\mathbf{V})$.

If we have access to interventional distributions $\mathbb{Z} \equiv \{p(\mathbf{V}_{i} (\mathbf{b}_{i} ))\}_{i=1}^k$ for $\mathbf{V}_{i}(\mathbf{b}_i ) \equiv (\mathbf{V} \setminus \mathbf{B}_i)(\mathbf{b}_i)$ instead of 
$p(\mathbf{V})$, then it is possible to identify distributions $p(\mathbf{Y}(\mathbf{a}))$ by checking whether
every district $\mathbf{D}$ in some SWIG ${\cal G}(\mathbf{Y}^*(\mathbf{a}))$ is in the set of intrinsic sets ${\cal I}({\cal G}(\mathbf{V}_{i}(\mathbf{b}_i)))$ for some $i \in 1, \ldots k$. It is possible that $\mathbf{D}$ is reachable in multiple interventional distributions $p(\mathbf{V}_{i}(\mathbf{b}_i))$ -- if so, we choose any one of the indices $i$ to assign to $\mathbf{D}$, henceforth denoted $i_{\mathbf{D}}$. The corresponding interventional distribution is denoted $p(\mathbf{V}_{i_{\mathbf{D}}} (\mathbf{b}_{i_{\mathbf{D}}}))$.
Since ${\bf D}\in {\cal I}({\cal G}(\mathbf{V}_{i}(\mathbf{b}_i)))$, 
$p(\mathbf{D}(\text{do}(\pas_{\cal G}(\mathbf{D}) )))$ may be obtained by a sequence of fixing operations on $p(\mathbf{V}_{i_{\mathbf{D}}} (\mathbf{b}_{i_{\mathbf{D}}}))$. 
This gives rise to the following result.

\begin{lem} \label{lem:gid}
    Fix a hidden variable causal model represented by an ADMG ${\cal G}(\mathbf{V})$, and a set of interventional distributions $\mathbb{Z} \equiv \{p(\mathbf{V}_i (\mathbf{b}_i))\}_{i=1}^k$  where $\mathbf{V}_i = \mathbf{V} \setminus \mathbf{B}_i$.   Then, $p(\mathbf{Y}(\mathbf{a}))$ is identified if and only if for each $\mathbf{D} \in {\cal D}({\cal G}(\mathbf{Y}^* (\mathbf{a})))$ where $\mathbf{Y}^* \equiv \an_{\mathcal{G}(\mathbf{V}(\mathbf{a}))}(\mathbf{Y}) \setminus \mathbf{a}$, there exists at least one $i_{\mathbf{D}} \in \{1, \ldots, k\}$ such that $p(\mathbf{V}_{i_{\mathbf{D}}}(\mathbf{b}_{i_{\mathbf{D}}})) \in \mathbb{Z}$ and $\mathbf{D} \in {\cal I}({\cal G} (\mathbf{V}_{i_{\mathbf{D}}} (\mathbf{b}_{i_{\mathbf{D}}})))$.
    Moreover, if $p(\mathbf{Y}(\mathbf{a}))$ is identified, it is equal to:
    \begin{align*}
        \sum_{\mathbf{Y}^* \setminus \mathbf{Y}}
        \!\!
        \prod_{\substack{\mathbf{D} \in \\{\cal D}({\cal G}({\mathbf{Y}^*}(\mathbf{a})))}}
        \!\!\!\!\!
        \!\!\!\!
        \!\!
        \phi_{\mathbf{V}_{i_{\mathbf{D}}} \setminus \mathbf{D}}(p(\mathbf{V}_{i_{\mathbf{D}}}(\mathbf{b}_{i_{\mathbf{D}}})); {\cal G}(\mathbf{V}_{i_{\mathbf{D}}}(\mathbf{b}_{i_{\mathbf{D}}}))) \vert_{\mathbf{A} = \mathbf{a}}.
    \end{align*}
\end{lem}

This formulation is equivalent to one given in \citet{leeGeneralIdentifiabilityArbitrary2019} (the proof appears in the Appendix).
An involved proof  \footnote{A minor edge case was not covered in this proof - namely the thicket construction is invalid for root sets with exactly one variable. We provide a correction in the Appendix.} in \citet{leeGeneralIdentifiabilityArbitrary2019} also shows gID is complete, meaning that if some $\mathbf{D}$ exists such that the fixing operator cannot be used to obtain the distribution
$p(\mathbf{D} | \text{do}(\pas_{\cal G}(\mathbf{D}) ))$ from any available interventional distribution and corresponding SWIG, then $p(\mathbf{Y}(\mathbf{a}))$ is not identified from those distributions.

Applying Lemma ~\ref{lem:gid} to identification of $p(Y(x_1, x_2))$ in Fig. ~\ref{fig:latentu1} with access to $\mathbb{Z} = \{ p(\{\mathbf{V} \setminus{X_1}\}(x_1)),  p(\{\mathbf{V} \setminus {X_2}\}(x_2))\}$ (represented by Figs. ~\ref{fig:latentu2} and ~\ref{fig:latentu3}) yields

        {\small\begin{align*}
        \sum_{W, U} p(Y(x_2)|  U(x_2), W) p(U(x_2))
        p(W(x_1) )
        \end{align*}}
The details are provided in the Appendix.
\subsection{aID: Identification with Ancestral Marginal Interventional Distributions}

The gID algorithm described above inherits the attractive feature of the ID algorithm that identification reduces to checking district pieces of a special set $\mathbf{Y}^*$ corresponding to causally relevant ancestors of $\mathbf{Y}$.
The limitation of gID is the requirement that interventional distribution inputs take the form of $p(\mathbf{V}_{i}(\mathbf{b}_i))$, where $\mathbf{V}_{i}$ and $\mathbf{B}_i$ are disjoint sets and their union yields $\mathbf{V}$.
In reality, as discussed in Section ~\ref{sec:motivating}, interventional distributions that are likely to be available will only be functions of $p(\mathbf{V}_{i}(\mathbf{b}_i))$ -- for example, a marginal distribution.

\begin{dfn} (Ancestrality)
Given a SWIG ${\cal G}(\mathbf{V}(\mathbf{a}))$, a set of random vertices $\mathbf{S}(\mathbf{a}) \subseteq \mathbf{V}(\mathbf{a})$ is said to be \emph{ancestral} if whenever $S(\mathbf{a}) \in \mathbf{S}(\mathbf{a})$, then $\an_{{\cal G}(\mathbf{V}(\mathbf{a}))}(S(\mathbf{a})) \setminus \mathbf{a} \subseteq \mathbf{S}(\mathbf{a})$.
\end{dfn}

We have the following result, which states that the gID algorithm may be adapted without loss of generality to the setting where all inputs are marginal interventional distributions with a particular property -- namely, that they are ancestral in their corresponding SWIG.
\begin{lem}\label{lem:aid}
    Fix a hidden variable causal model represented by an ADMG ${\cal G}(\mathbf{V})$, and a set of interventional distributions
    $\mathbb{Z} \equiv \{p(\mathbf{S}_{i} (\mathbf{b}_{i} ))\}_{i=1}^k$,
    such that each $\mathbf{S}_i(\mathbf{b}_i)$ is ancestral in ${\cal G}(\mathbf{V}({\bf b}_i))$.  Then $p(\mathbf{Y}(\mathbf{a}))$ is identified if and only if for each $\mathbf{D} \in {\cal D}({\cal G}(\mathbf{Y}^*(\mathbf{a})))$ where $\mathbf{Y}^* \equiv \an_{{\cal G}(\mathbf{V}(\mathbf{a}))}(\mathbf{Y}) \setminus \mathbf{a}$, there exists at least one $i_{\mathbf{D}} \in \{ 1, \ldots, k \}$ such that $p(\mathbf{S}_{i_{\mathbf{D}}}(\mathbf{b}_{i_{\mathbf{D}}})) \in {\mathbb Z}$ and
    $\mathbf{D} \in {\cal I}({\cal G}({\mathbf S}_{i_{\mathbf{D}}}(\mathbf{b}_{i_{\mathbf{D}}})))$.
    Moreover, if $p(\mathbf{Y}(\mathbf{a}))$ is identified, it is equal to:
    \begin{align*}
        \sum_{\mathbf{Y}^* \setminus \mathbf{Y}} \prod_{\mathbf{D} \in {\cal D}({\cal G}({\mathbf{Y}^*}(\mathbf{a})))}
        \!\!\!\!\!\!\!\!
        \!\!\!\!\!
        \phi_{\mathbf{S}_{i_{\mathbf{D}}} \setminus \mathbf{D}}(p(\mathbf{S}_{i_{\mathbf{D}}}(\mathbf{b}_{i_{\mathbf{D}}})); {\cal G}(\mathbf{S}_{i_{\mathbf{D}}}(\mathbf{b}_{i_{\mathbf{D}}}))) \vert_{\mathbf{A} = \mathbf{a}}.
    \end{align*}
\end{lem}

Applying Lemma \ref{lem:aid} to identification of $p(Y(x_1, x_2))$ in Fig.~\ref{fig:latentu1} with 
$\mathbb{Z}= \{p(W(x_1)), 
p(\{W, U, Y, X_1\}(x_2)) \}$ (represented by Figs.~\ref{fig:latentu4} and ~\ref{fig:latentu3}) yields
{\small
    \begin{align*}
        \sum_{W, U} p(Y(x_2) \mid W, U(x_2), X_1) p(U(x_2)\mid X_1) \times p(W(x_1)).
    \end{align*}
}
The details are provided in the Appendix.

In the remainder of the paper, we consider increasingly general identifications algorithms for $p(\mathbf{Y}(\mathbf{a}))$ that allow arbitrary marginal distributions obtained from $p(\mathbf{V}_{i}(\mathbf{b}_i))$, and then conditionals distributions obtained from $p(\mathbf{V}_{i}(\mathbf{b}_i))$ to be used as inputs.
\subsection{mID: Identification with Marginal Interventional Distributions}

If marginal interventional distributions given as input are not ancestral in the corresponding SWIG, the identification algorithm becomes considerably more complicated.  In particular, it is no longer sufficient to consider the set $\mathbf{Y}^* \equiv \an_{{\cal G}(\mathbf{V}(\mathbf{a}))}(\mathbf{Y}) \setminus \mathbf{a}$.  Consider the following example.

In Fig.~\ref{fig:latentu1} we wish to identify $p(Y(x_1, x_2))$.  If we consider the set $Y^* = \{ U, Y, W \}$ to try to identify this distribution, we will conclude that the required intrinsic sets are $\mathcal{D}(\mathcal{G}_{\mathbf{Y}^*}) = \{\{ U, Y \}, \{ W\}\}$, which means that we must identify the corresponding interventional distributions $p(\{U, Y\}(w, x_2)), p(W(x_1))$.   However, assume that we only have access to the (non-ancestral) marginal interventional distributions $\mathbb{Z} = \{ p(W(x_1)), p(Y(x_2), W(x_2))\}$, represented by Figs.~\ref{fig:latentu4} and ~\ref{fig:latentu6} respectively.
In this case, using the aID algorithm below will fail to identify $p(Y(x_1, x_2))$ since $p(\{U, Y\}(w, x_2))$ is 
not identifiable from any distribution in $\mathbb{Z}$, since none of them have any information on $U$.

However, it is possible to identify $p(Y(x_1, x_2))$ from $\mathbb{Z}$ via a larger set $Y'$ that contains $Y$ but not $U$ (despite the fact that $U(x_1,x_2)$ is an ancestor of $Y(x_1,x_2)$ in the SWIG ${\cal G}(\mathbf{V}(x_1,x_2))$).
Specifically, let $\mathbf{Y}'  \equiv \an_{\mathcal{G}((\mathbf{V} \setminus \{ U \})(\mathbf{a}))} (\mathbf{Y}) \setminus \mathbf{a} = \{ Y, W \}$. Then, $\mathcal{D}(\mathcal{G}_{\mathbf{Y}'}) = \{\{Y\}, \{W \}\}$.
It is easy to verify that $\{W\} \in \mathcal{I}(\mathcal{G}(W(x_1)))$, and $\{Y\} \in \mathcal{I}(\mathcal{G}(Y (w, x_2)))$, and thus $p(Y(x_1, x_2))$ is identified from $\mathbb{Z}$.

The algorithm schema we present here considers all possible subsets $\mathbf{Y}'$ of $\mathbf{Y}^*$ that also include $\mathbf{Y}$.
\begin{lem} \label{lem:mid}
    Given a hidden variable causal model represented by an ADMG ${\cal G}(\mathbf{V})$, and a set of interventional distributions
    $\mathbb{Z} \equiv \{p(\mathbf{S}_{i} (\mathbf{b}_{i} ))\}_{i=1}^k$,
    then $p(\mathbf{Y}(\mathbf{a}))$ is identified from $\mathbb{Z}$ if there exists $\mathbf{Y}' \subseteq {\cal P}(\mathbf{Y}^* \setminus \mathbf{Y}) \cup \mathbf{Y}$,\footnote{${\cal P}(\mathbf{S})$ for any set $\mathbf{S}$ is the power set of $\mathbf{S}$.} such that for each $\mathbf{D} \in {\cal D}({\cal G}(\mathbf{Y}'(\mathbf{a})))$, we can find at least one $i_{\mathbf{D}}$ such that $p(\mathbf{S}_{i_{\mathbf{D}}}(\mathbf{b}_{i_{\mathbf{D}}})) \in {\mathbb Z}$,
    $\pas_{\mathcal{G}(\mathbf{S}_{i_{\mathbf{D}}}(\mathbf{b}_{i_{\mathbf{D}}}))} (\mathbf{D}) = \pas_{\mathcal{G}(\mathbf{Y}')}(\mathbf{D})$, and $\mathbf{D} \in {\cal I}({\cal G}(\mathbf{S}_{i_{\mathbf{D}}} (\mathbf{b}_{i_{\mathbf{D}}})))$.
    Moreover, if $p(\mathbf{Y}(\mathbf{a}))$ is identified, the identifying formula is:
    \begin{align*}
        \sum_{\mathbf{Y}' \setminus \mathbf{Y}} \prod_{\mathbf{D} \in {\cal D}({\cal G}({\mathbf{Y}'}(\mathbf{a})))}
        \!\!\!\!\!\!\!\!
        \!\!\!\!
        \phi_{\mathbf{S}_{i_{\mathbf{D}}} \setminus \mathbf{D}}(p(\mathbf{S}_{i_{\mathbf{D}}}(\mathbf{b}_{i_{\mathbf{D}}})); {\cal G}(\mathbf{S}_{i_{\mathbf{D}}}(\mathbf{b}_{i_{\mathbf{D}}}))) \vert_{\mathbf{A} = \mathbf{a}}.
    \end{align*}

\end{lem}

Note that evaluating whether an appropriate $\mathbf{Y}'$ exists is intractable in general.  We emphasize this point by calling this an algorithm schema rather than a tractable algorithm.  A polynomial time check that would discover an appropriate $\mathbf{Y}'$, if it exists, is currently an open question.

Applying Lemma ~\ref{lem:mid} to identifying $p(Y(x_1, x_2))$ with $\mathbb{Z} = \{p(W(x_1)), p(\{Y, W\}(x_2))\}$ yields
$\sum_{W} p(Y(x_2)\mid W) p(W(x_1)).$
The details are provided in the Appendix.

\subsection{eID: Identification with Arbitrary Conditional Interventional Distributions}
We now consider the case where interventional distributions might arise as arbitrary marginal or conditional distributions. Conditional distributions can arise if data were collected on a subset of a population (e.g. an RCT is conducted with specific enrollment criteria).  We consider any combination of distributions of the form $p({\bf S}_i({\bf b}_i) | {\bf C}_i({\bf b}_i))$ (available at all levels of ${\bf C}_i({\bf b}_i)$), and distributions of the form $p({\bf S}_i({\bf b}_i) | {\bf C}_i({\bf b}_i) = {\bf c}_i)$ (available only at a specific set of values ${\bf c}_i$ of ${\bf C}_i({\bf b}_i)$).

Considering interventional conditional distributions creates additional complications.  First, identification may have to proceed in an interventional distribution where some variables are also conditioned on.  We adapt the algorithm in \citet{bareinboim15selection} for this task, rephrasing it in terms of intrinsic sets and a modification of the fixing operator (adapted from \citet{bhattacharyaIdentificationMissingData2019}).  Second, identification of an interventional distribution from conditionals may in general require us to ``stitch together''  multiple distributions via the chain rule of probability.  We address this issue by a preprocessing step applied to the input set $\mathbb{Z}$ that uses the chain rule and model restrictions to construct all additional distributions not already present in $\mathbb{Z}$.

\begin{dfn}
    A variable $A \subseteq \mathbf{V}$ is \emph{selection fixable (s-fixable)} in a CADMG ${\cal G}(\mathbf{V},\mathbf{W})$ given conditioned variables $\mathbf{C} \subseteq \mathbf{V}$ if
    $\mathbf{C} \cap \de_{{\cal G}(\mathbf{V},\mathbf{W})}(A) = \emptyset$ and $\de_{{\cal G}(\mathbf{V},\mathbf{W})}(A) \cap \dis_{{\cal G}(\mathbf{V},\mathbf{W})}(A) = \{ A \}$.
\end{dfn}

\begin{dfn}
    If $A \subseteq \mathbf{V}$ is s-fixable in a CADMG ${\cal G}(\mathbf{V},\mathbf{W})$ where $\mathbf{C}\subseteq\mathbf{V}$ is conditioned, define the s-fixing operator $\phi^{\mathbf{C}}_A({\cal G}(\mathbf{V},\mathbf{W}))$ as
    $\phi_A({\cal G}(\mathbf{V},\mathbf{W}))$ (the ordinary fixing operator on graphs), yielding ${\cal G}(\mathbf{V} \setminus \{ A \}, \mathbf{W} \cup \{ A \})$.
\end{dfn}
Note that if $\mathbf{C}\subseteq\mathbf{V}$ is conditioned in 
${\cal G}(\mathbf{V},\mathbf{W})$, $\mathbf{C}\subseteq\mathbf{V}\setminus \{A\}$ is conditioned in $\phi^{\mathbf{C}}_A({\cal G}(\mathbf{V},\mathbf{W}))$.

\begin{dfn}
    If $A \subseteq \mathbf{V}$ is s-fixable in a CADMG ${\cal G}(\mathbf{V},\mathbf{W})$ where $\mathbf{C}\subseteq\mathbf{V}$ is conditioned, associated with a kernel $q_{\mathbf V}(\mathbf{V} \setminus \mathbf{C} \mid \mathbf{W} = \mathbf{w}, \mathbf{C} = \mathbf{c})$, define the s-fixing operator on kernels as:
    \begin{align*}
&\phi^{\mathbf{C}}_{A}(q_{\mathbf V}(\mathbf{V} \setminus \mathbf{C} \mid \mathbf{W}, \mathbf{C} = \mathbf{c}), {\cal G}(\mathbf{V},\mathbf{W})) \\
&\equiv \frac{q_{\mathbf V}(\mathbf{V} \setminus \mathbf{C} \mid \mathbf{W} = \mathbf{w}, \mathbf{C} = \mathbf{c})}{q_{\mathbf V}(A \mid \mb_{{\cal G}(\mathbf{V},\mathbf{W})}(A) \setminus \mathbf{C}, \mathbf{C} = \mathbf{c}, \mathbf{W} = \mathbf{w})}
    \end{align*}
\end{dfn}

The s-fixing operation of $A$ given a conditioned set $\mathbf{C}$, and the conditioning operation on $\mathbf{C}$ commute, in the following sense.
\begin{lem}\label{lem:commute}
    If $A \subseteq \mathbf{V}$ is s-fixable in a CADMG ${\cal G}(\mathbf{V},\mathbf{W})$ where $\mathbf{C}\subseteq\mathbf{V}$ is conditioned, associated with a kernel $q_{\mathbf V}(\mathbf{V} \setminus \mathbf{C} \mid \mathbf{W} = \mathbf{w}, \mathbf{C} = \mathbf{c})$, then
    \begin{align*}
&q_{\mathbf{V}\setminus\{A\}}(\mathbf{V}\setminus(\mathbf{C}\cup\{A\})|\mathbf{W}\cup\{A\}, \mathbf{C}=\mathbf{c}) =\\
&\qquad \phi^{\mathbf{C}}_A(q_{\mathbf{V}}(\mathbf{V}\setminus\mathbf{C} | \mathbf{W},\mathbf{C}=\mathbf{c}), {\cal G}(\mathbf{V},\mathbf{W}))\text{, with}\\
& q_{\mathbf{V}\setminus\{A\}}(\mathbf{V}\setminus\{A\}|\mathbf{W}\cup\{A\}) \equiv \phi_A(q_{\mathbf{V}}(\mathbf{V} | \mathbf{W}), {\cal G}(\mathbf{V},\mathbf{W})).
    \end{align*}
\end{lem}
\begin{dfn}
    A sequence $\sigma_{\mathbf{A}}$ of elements in $\mathbf{A}$ is s-fixable in ${\cal G}(\mathbf{V},\mathbf{W})$ given
    a conditioned $\mathbf{C} \subseteq \mathbf{V}$ if
    either $\mathbf{A} = \emptyset$, or $\eta(\sigma_{\mathbf{A}})$ is s-fixable in ${\cal G}(\mathbf{V},\mathbf{W})$, and
    $\tau(\sigma_{\mathbf{A}})$ is s-fixable in $\phi^{\mathbf{C}}_{\eta(\sigma_{\mathbf{A}})}({\cal G}(\mathbf{V},\mathbf{W}))$.
\end{dfn}
$\mathbf{A}$ is s-fixable in ${\cal G}(\mathbf{V},\mathbf{W})$ if there exists an s-fixable sequence $\sigma_{\mathbf{A}}$ in ${\cal G}(\mathbf{V},\mathbf{W})$.

The following is a version of the ID algorithm when the input distribution has selection  (conditioning on a particular value), in terms of the s-fixing operator.
\begin{lem}
    Given a SWIG ${\cal G}(\mathbf{S}(\mathbf{b}))$, and the corresponding interventional distribution $p(\mathbf{S}(\mathbf{b}))$,
    let $\mathbf{Y}^* = (\an_{{\cal G}(\mathbf{S}(\mathbf{b}))}(\mathbf{Y}) \setminus \mathbf{a})$. Let $\mathbf{Y} \subseteq \mathbf{S}$,
    and $\mathbf{A} \subseteq \mathbf{S} \cup \mathbf{B}$.
    $p(\mathbf{Y}(\mathbf{a}))$ is identified from $p(\{\mathbf{S} \setminus \mathbf{C}\}(\mathbf{b}) | \mathbf{C}(\mathbf{b}) = \mathbf{c})$
    if $\de_{{\cal G}({\bf S}({\bf b}))}(\mathbf{Y}^*) \cap \mathbf{C} = \emptyset$,
    $\mathbf{c}_{\pa_{\cal G}(\mathbf{Y}^*) \cap \mathbf{A}}$ is consistent with $\mathbf{a}$,
    $\mathbf{b}_{\pa_{\cal G}(\mathbf{Y}^*) \cap \mathbf{A}}$ is consistent with $\mathbf{a}$,
    and for each district $\mathbf{D} \in {\cal D}({\cal G}(\mathbf{Y}^*(\mathbf{a})))$,
    there exists a set ${\bf Z}_{\bf D} \in {\bf S} \setminus {\bf C}$, such that
    ${\bf D} \subseteq {\bf Z}_{\bf D}$,
    ${\bf Z}_{\bf D}$ is s-fixable in ${\cal G}(\mathbf{S}(\mathbf{b}))$ (given a conditioned ${\bf C}$) by a sequence
    $\sigma_{{\bf Z}_{\bf D}}$ that fixes $\bar{\bf D} \subseteq {\bf Z}_{\bf D}$ last, where $\bar{\bf D}$ is a district in $\phi^{\bf C}_{\sigma_{{\bf Z}_{\bf D} \setminus \bar{\bf D}}}({\cal G}({\bf S}({\bf b})))$, and
    ${\bf D}$ is reachable in $\phi^{\bf C}_{\sigma_{{\bf Z}_{\bf D} \setminus \bar{\bf D}}}({\cal G}({\bf S}({\bf b})))$.

    If $p(\mathbf{Y}(\mathbf{a}))$ is identified, it is equal to
    \begin{align*}
        \sum_{\mathbf{Y}^* \setminus \mathbf{Y}} \prod_{\mathbf{D} \in {\cal D}({\cal G}(\mathbf{Y}^*({\mathbf a})))} q_{\bf D}({\bf D} | \pa^s_{{\cal G}({\bf S}({\bf b}))}({\bf D})) \vert_{{\bf A}={\bf a}},
    \end{align*}
    where for each ${\bf D} \in {\cal D}({\cal G}(\mathbf{Y}^*({\mathbf a})))$,
    {\small
        \begin{align*}
            &q_{\bf D}\!({\bf D} |\! \pa^s_{{\cal G}({\bf S}({\bf b}))}\!({\bf D}))\! \equiv \!\phi_{\bar{\bf D} \setminus {\bf D}}\!(q_{\bar{\bf D}}(\bar{\bf D} | \!\pa^s_{{\cal G}}\!(\bar{\bf D}));\!
            \phi_{{\bf S}\! \setminus \!\bar{\bf D}}({\cal G}({\bf S}({\bf b}))))\\
            &q_{\bar{\bf D}}(\bar{\bf D} | \pa^s_{{\cal G}}(\bar{\bf D})) \equiv \prod_{D \in \bar{\bf D}} q_{{\bf S} \setminus ({\bf Z}_{\bf D} \setminus \bar{\bf D})}(D | \mb^*(D),
            {\bf Z}_{\bf D} \setminus \bar{\bf D})\\
            &q_{{\bf S} \setminus ({\bf Z}_{\bf D} \!\setminus\! \bar{\bf D})}({\bf S} \!\setminus\! ({\bf Z}_{\bf D} \!\setminus\! \bar{\bf D}) | {\bf Z}_{\bf D} \!\setminus\! \bar{\bf D}) \equiv \\
            & \quad\phi^{\bf C}_{{\bf Z}_{\bf D} \!\setminus\! \bar{\bf D}}(p(\{ {\bf S} \setminus {\bf C} \}({\bf b}) | {\bf C}({\bf b}) = {\bf c}; {\cal G}({\bf S}({\bf b})))),
        \end{align*}
    }
    with $\mb^*(D)$ defined as $\mb_{\phi_{{\bf Z}_{\bf D} \setminus \bar{\bf D}}({\cal G}({\bf S}({\bf b})))}(D)$ intersected with elements in $\bar{\bf D}$ earlier than $D$ in any reverse topological order in $\phi_{{\bf Z}_{\bf D} \setminus \bar{\bf D}}({\cal G}({\bf S}({\bf b})))$.
    \label{lem:joint-from-conditional}
\end{lem}

\begin{dfn}
    The set $\mathbb{Z}$ 
    is said to be \emph{chain rule closed} if for any
    $p(\mathbf{S}_i(\mathbf{b}_i) | \mathbf{C}_i(\mathbf{b}_i)) \in \mathbb{Z}$, and a partition $\mathbf{C}^1_i(\mathbf{b}_i) \dot{\cup} \mathbf{C}^2_i(\mathbf{b}_i)$ of $\mathbf{C}_i(\mathbf{b}_i)$, if there exists
    $p(\mathbf{S}_j(\mathbf{b}_j) | \mathbf{C}_j(\mathbf{b}_j)) \in \mathbb{Z}$ such that
    $p(\mathbf{C}^1_i(\mathbf{b}_i) | \mathbf{C}^2_i(\mathbf{b}_i)) = p(C_i^1(\mathbf{b}_j) | \mathbf{S}_j(\mathbf{b}_j) \setminus C_i^1(\mathbf{b}_j), C_j(\mathbf{b}_j))$,
    under the given causal model,
    then $p(\mathbf{S}_i(\mathbf{b}_i), \mathbf{C}^1_i(\mathbf{b}) | \mathbf{C}_i(\mathbf{b}_i) \setminus \mathbf{C}^1_i(\mathbf{b}_i)) \in \mathbb{Z}$.
\end{dfn}
Any set of conditional counterfactual distributions $\mathbb{Z}$ may also be made chain rule closed without loss of generality, and the required equality may be established by rules of po-calculus in \cite{malinsky19po}.

\begin{lem}\label{lem:eid_statement}
    Fix a hidden variable causal model represented by an ADMG ${\cal G}(\mathbf{V})$, and a chain rule closed set of distributions ${\mathbb Z} = \{ p(\mathbf{S}_i(\mathbf{b}_i) | \mathbf{C}_i(\mathbf{b}_i)) \}_{i=1}^k$ (with some possibly available only at a level ${\bf c}_i$).
    Then $p(\mathbf{Y}(\mathbf{a}))$ is identified from $\mathbb{Z}$ if
    there exists $\mathbf{Y}' \subseteq {\cal P}(\mathbf{Y}^* \setminus \mathbf{Y}) \cup \mathbf{Y}$, such that for
    for each $\mathbf{D} \in {\cal D}({\cal G}(\mathbf{Y}'(\mathbf{a})))$, we can find at least one $i_{\mathbf{D}}$ such that
    $\pas_{\mathcal{G}(\{ \mathbf{S}_{i_{\mathbf{D}}} \cup \mathbf{C}_{i_{\mathbf{D}}} \}(\mathbf{b}_{i_{\mathbf{D}}}) )} (\mathbf{D}) = \pas_{\mathcal{G}(\mathbf{Y}')}(\mathbf{D})$,
    and
    $p(\mathbf{D} \mid \doo(\pas_{\mathcal{G}(\mathbf{S}_{i_{\mathbf{D}}}(\mathbf{b}_{i_{\mathbf{D}}}) \cup \mathbf{C}_{i_{\mathbf{D}}}(\mathbf{b}_{i_{\mathbf{D}}}))} (\mathbf{D})))$ is identified from $p(\mathbf{S}_i(\mathbf{b}_i) | \mathbf{C}_i(\mathbf{b}_i))$ evaluated at ${\bf c}_i$ consistent with ${\bf a}$ using Lemma \ref{lem:joint-from-conditional}.
    Moreover, if $p(\mathbf{Y}(\mathbf{a}))$ is identified, it is equal to:
    \begin{align*}
        \sum_{\mathbf{Y}' \setminus \mathbf{Y}} \prod_{\mathbf{D} \in {\cal D}({\cal G}({\mathbf{Y}'}(\mathbf{a})))}
        \!\!\!\!\!\!\!\!
        p(\mathbf{D}
        (\pas_{\mathcal{G}(\mathbf{S}_{i_{\mathbf{D}}}(\mathbf{b}_{i_{\mathbf{D}}}) \cup \mathbf{C}_{i_{\mathbf{D}}}(\mathbf{b}_{i_{\mathbf{D}}}))} (\mathbf{D}))) \vert_{\mathbf{A}=\mathbf{a}} \\
        =
        \sum_{\mathbf{Y}' \setminus \mathbf{Y}} \prod_{\mathbf{D} \in {\cal D}({\cal G}({\mathbf{Y}'}(\mathbf{a})))}
        q_{\mathbf{D}}(\mathbf{D} | \pas_{{\cal G}_{\mathbf{Y}'}}(\mathbf{D}) ) \vert_{\mathbf{A}=\mathbf{a}},
    \end{align*}
    where each $q_{\mathbf{D}}(\mathbf{D} | \pas_{{\cal G}_{\mathbf{Y}'}}(\mathbf{D}) )$ is obtained from applying Lemma ~\ref{lem:joint-from-conditional} to the appropriate element of $\mathbb{Z}$.
\end{lem}

A worked example of Lemma ~\ref{lem:eid_statement} is provided in the Appendix.

\section{Completeness}
\label{sec:complete}
An identification algorithm is considered \textit{complete} if it fails only when no computable functional exists.

We consider a proof of completeness for aID. The aim is to demonstrate that aID will only fail when there exists a structure in the graph that inhibits identification, by allowing the creation of two models $\mathcal{M}_1, \mathcal{M}_2$ which agree on the input distributions, but disagree on a causal effect.

For disjoint sets $\mathbf{A}, \mathbf{Y}$, the causal effect $p(\mathbf{Y}(\mathbf{a}))$ is not identified from a set of ancestral marginal distributions $\mathbb{Z}=\{p(\mathbf{S}_i (\mathbf{b}_i))\}_{i=1}^k$ if there exist distinct causal models ${\cal M}_1, {\cal M}_2$ such that $p^1 (\mathbf{S}_i (\mathbf{b}_i)) = p^2(\mathbf{S}_i ({\mathbf b}_i))$ for all $i \in \{1, \ldots, k\}$, but $p^1 (\mathbf{Y}(\mathbf{a})) \neq p^2 (\mathbf{Y} (\mathbf{a}))$.

We consider a set of distributions $\bar{\mathbb Z} = \{p(\mathbf{V}_i (\mathbf{b}_i))\}_{i=1}^k$, where the interventions $\mathbf{b}_i$ are identical to those in $\mathbb{Z}$. Precisely, we construct arbitrary $p(\{\mathbf{V}_i \setminus \mathbf{S} \}(\mathbf{b}_i) \mid \mathbf{S}_i (\mathbf{b}_i))$ for $i = 1, \ldots, k$. These distributions are combined with models ${\cal M}_1, {\cal M}_2$ as
\begin{align*}
    p^1(\mathbf{V}_i (\mathbf{b}_i)) &= p^1 (\mathbf{S}_i (\mathbf{b}_i))p(\{\mathbf{V}_i \setminus \mathbf{S} \}(\mathbf{b}_i) \mid \mathbf{S}_i (\mathbf{b}_i)), \\
    p^2(\mathbf{V}_i (\mathbf{b}_i)) &= p^2 (\mathbf{S}_i (\mathbf{b}_i))p(\{\mathbf{V}_i \setminus \mathbf{S} \}(\mathbf{b}_i) \mid \mathbf{S}_i (\mathbf{b}_i)).
\end{align*}

This construction is the input to gID - interventional distributions over $\mathbf{V}$. Logically, if a causal query $p(\mathbf{Y}(\mathbf{a}))$ fails against $\mathbb{Z}$ using aID, it can either fail with $\bar{\mathbb{Z}}$ using gID, or succeed with gID. In the former case, the thicket construction proving non-identification in gID can be adapted by marginalization to prove non-identification in aID (proved in Lemma ~\ref{lem:gid_fails}). In the latter case, we are required to demonstrate that the failure of aID was due to some graphical object preventing identification. Due to the ancestrality property of distributions in $\mathbb{Z}$, it happens that this object is also a thicket (proved in Lemma ~\ref{lem:aid_complete}).

\begin{lem}\label{lem:gid_fails}
    If a causal query $p(\mathbf{Y}(\mathbf{a}))$ fails from $\mathbb{Z}$ using aID, and fails from $\bar{\mathbb{Z}}$ using gID, then this causal query is not identified.
\end{lem}

\begin{lem}\label{lem:aid_fails}

    If a causal query $p(\mathbf{Y}(\mathbf{a}))$ fails from $\mathbb{Z}$ using aID, but succeeds from $\bar{\mathbb Z}$ using gID, then a thicket construction demonstrating non-identifiability applies.
\end{lem}

The above results taken together establish completeness of aID.

\begin{thm}\label{lem:aid_complete}
    aID is complete.

\end{thm}

\section{Conclusions}
\label{sec:conclusion}

In this paper we used Single World Intervention Graphs (SWIGs) \cite{thomas13swig}, the potential outcomes calculus \cite{malinsky19po}, and the nested Markov factorization of mixed graphs \cite{richardson17nested} to yield a set of increasingly general algorithms for identification of counterfactual distributions given an arbitrary set of counterfactual or observed data distributions as inputs.
These results generalize a previous algorithm described in \cite{leeGeneralIdentifiabilityArbitrary2019}.
In addition, we show that for a class of marginal counterfactual distribution inputs, our algorithm is complete.

Since our algorithm formulation relies on the nested Markov factorization of mixed graphs, it naturally lends itself to parametric statistical inference for cases where nested Markov likelihoods have been formulated, such as discrete and Gaussian data.  Giving estimators for functionals identified by our algorithms for likelihoods for more general types of data, as well as deriving semi-parametric estimators \cite{tsiatis06missing} are obvious areas of future work.

In addition, important open problems include showing whether all algorithms we describe are complete, as well as developing efficient implementations in software.

\bibliography{references}
\bibliographystyle{icml2020}

\appendix

\end{document}


\onecolumn\icmltitle{Appendix to Identification Methods With Arbitrary Interventional Distributions as Inputs}




\begin{icmlauthorlist}
\icmlauthor{Jaron J. R. Lee}{jhu}
\icmlauthor{Ilya Shpitser}{jhu}
\end{icmlauthorlist}

\icmlaffiliation{jhu}{Department of Computer Science, Johns Hopkins University, Baltimore, Maryland, USA}

\icmlcorrespondingauthor{Jaron J. R. Lee}{jaron.lee@jhu.edu}

\icmlkeywords{Machine Learning, ICML}

\vskip 0.3in




\section{Proofs of Soundness}\label{sec:sound}
We begin by reproducing the proof of soundness of the ID algorithm in \citet{richardsonNestedMarkovProperties2017}, which will serve as a framework for an alternative proof of soundness for gID and 
other soundness proofs of generalizations of gID we described.
\begin{thm}\label{thm:id}
    Let $\mathcal{G}(\mathbf{H} \cup \mathbf{V})$ be a causal DAG with a latent projection $\mathcal{G}(\mathbf{V})$. For $\mathbf{A} \dot{\cup} \mathbf{Y} \subset \mathbf{V}$, let $\mathbf{Y}^* = \an_{\mathcal{G}(\mathbf{V}(\mathbf{a}))} (\mathbf{Y}) \setminus a$. If $\mathcal{D}(\mathcal{G}(\mathbf{Y}^*(\mathbf{a})) ) \subseteq \mathcal{I}(\mathcal{G}(\mathbf{V})) $ then $p(\mathbf{Y}(\mathbf{a}))$ is identified and given by 
    \[
    \sum_{\mathbf{Y}^* \setminus \mathbf{Y}} \prod_{\mathbf{D} \in \mathcal{D}(\mathcal{G}(\mathbf{Y}^*(\mathbf{a})))} \phi_{\mathbf{V} \setminus \mathbf{D}} (p(\mathbf{V}); {\cal G}(\mathbf{V})) \vert_{\mathbf{A}=\mathbf{a}}.
    \]
\end{thm}
\begin{prf}
    Let $\mathbf{A}^* = \mathbf{V} \setminus \mathbf{Y}^* \supseteq \mathbf{A}$. 
    By Lemma 53 in \cite{richardsonNestedMarkovProperties2017}, $p(\mathbf{Y}^* \mid \doo(\mathbf{A})) = p(\mathbf{Y}^* \mid \doo(\mathbf{A}^*))$.

    Since $\mathcal{G}(\mathbf{H} \cup \mathbf{V}))$ is a DAG, any set is fixable, in particular $\mathbf{A}^*$.
    Let $\mathcal{G}^*(\mathbf{H} \cup (\mathbf{V} \setminus \mathbf{A}^*), \mathbf{A}^*)$ be the CDAG obtained from $\phi_{\mathbf{A}^*}(\mathcal{G}(\mathbf{H} \cup \mathbf{V}))$.

	Given a graph ${\cal G}(\mathbf{V}\cup\mathbf{H})$, define $\sigma_{\mathbf{H}}$ to be the operator that creates a latent projection that removes vertices in $\mathbf{H}$.
	By Corollary 53 in \cite{richardsonNestedMarkovProperties2017}, fixing ($\phi$) and latent projections($\sigma$) operators in any CDAG commute, in other words,
	$\sigma_{\mathbf{H}}(\phi_{\mathbf{A}^*}(\mathcal{G}(\mathbf{H} \cup \mathbf{V}))) = \phi_{\mathbf{A}^*}(\sigma_{\mathbf{H}}(\mathcal{G}(\mathbf{H} \cup \mathbf{V})))$.
    
    Since $\mathcal{G}^*(\mathbf{Y}^*, \mathbf{A}^*) = \sigma_{\mathbf{H}}(\phi_{\mathbf{A}^*}(\mathcal{G}(\mathbf{H} \cup \mathbf{V})))$, and by definition of induced subgraphs, $\mathcal{G}(\mathbf{V})_{\mathbf{Y}^*} = (\phi_{\mathbf{A}^*}(\mathcal{G}(\mathbf{V})))_{\mathbf{Y}^*}$, this means that $\mathcal{G}(\mathbf{V})_{\mathbf{Y}^*} = \mathcal{G}^*(\mathbf{Y}^*, \mathbf{A}^*)_{\mathbf{Y}^*}$. Therefore $\mathcal{D}(\mathcal{G}(\mathbf{V})_{\mathbf{Y}^*}) = \mathcal{D}(\mathcal{G}(\mathbf{Y}^*, \mathbf{A}^*))$.  Further, by definition of SWIG latent projections and induced subgraphs, $\mathcal{D}(\mathcal{G}(\mathbf{V})_{\mathbf{Y}^*}) = \mathcal{D}(\mathcal{G}({\mathbf{Y}^*}(\mathbf{a})))$. 

    For every district $\mathbf{D} \in \mathcal{D}(\mathcal{G}^*(\mathbf{Y}^*, \mathbf{A}^*))$, define $\mathbf{H}_{\mathbf{D}} \equiv \mathbf{H} \cap \an_{\mathcal{G}(\mathbf{H} \cup \mathbf{V})_{\mathbf{D} \cup \mathbf{H}}}(\mathbf{D}))$, and $\mathbf{H}^* = \bigcup_{\mathbf{D} \in \mathcal{D}(\mathcal{G}^*(\mathbf{Y}^*, \mathbf{A}^*)) }\mathbf{H}_{\mathbf{D}}$. $\mathbf{H}_{\mathbf{D}}$ is the set of variables $h \in \mathbf{H}$ for which there exists a vertex $d \in \mathbf{D}$ and a directed path $h \to \ldots \to d$ in $\mathcal{G}(\mathbf{H} \cup \mathbf{V})$ on which excepting $d$ all vertices are in $\mathbf{H}$.

    The construction of $\mathbf{H}_{\mathbf{D}}$ implies the following three corollaries.

    \begin{cor}\label{cor:a}
        If $\mathbf{D}, \mathbf{D}' \in \mathcal{D}(\mathcal{G}^*(\mathbf{Y}^*, \mathbf{A}^*))$ and $\mathbf{D} \neq \mathbf{D}'$, then $\mathbf{H}_{\mathbf{D}} \cap \mathbf{D}_{\mathbf{D}'} = \emptyset$.
    \end{cor}
    \begin{prf}
        This is because if the intersection was not empty, the two districts $\mathbf{D}, \mathbf{D}'$ would have a single $H \in \mathbf{H}$ pointing into both of them.
    \end{prf}
    \begin{cor}\label{cor:b}
        For each $\mathbf{D} \in \mathcal{D}(\mathcal{G}^*(\mathbf{Y}^*, \mathbf{A}^*))$ we have $\pas_{\mathcal{G}(\mathbf{H} \cup \mathbf{V})}(\mathbf{D} \cup \mathbf{H}_{\mathbf{D}}) \cap \mathbf{H}^* = \mathbf{H}_{\mathbf{D}}$.
    \end{cor}
    \begin{cor}\label{cor:c}
        $\mathbf{Y}^* \cup \mathbf{H}^*$ is ancestral in $\mathcal{G}(\mathbf{H} \cup \mathbf{V})$ so if $V \in \mathbf{Y}^* \cup \mathbf{H}^*$, $\pas_{\mathcal{G}(\mathbf{H} \cup \mathbf{V})}(V) \cap \mathbf{H} \subseteq \mathbf{H}^*$.
    \end{cor}

    Then, 

    \begin{align*}
        &p(\mathbf{Y}^* \mid \doo_{\mathcal{G}(\mathbf{H} \cup \mathbf{V})(\mathbf{A}^*)}) \\
        &= \sum_{\mathbf{H}} \prod_{V \in \mathbf{H} \cup \mathbf{Y}^*} p(V\mid \pas_{\mathbf{G}(\mathbf{H} \cup \mathbf{V})} (V))\\
        &= \sum_{\mathbf{H}^*} \prod_{V \in \mathbf{H}^* \cup \mathbf{Y}^*} p(V \mid \pas_{\mathcal{G}(\mathbf{H} \cup \mathbf{V})}(V)) \times \underbrace{\sum_{\mathbf{H}\setminus \mathbf{H}^*} \prod_{V \in \mathbf{H} \setminus \mathbf{H}^*} p(V \mid \pas_{\mathcal{G}(\mathbf{H} \cup \mathbf{V})}(V)) }_{=1}\\
        &=\sum_{\mathbf{H}^*} \prod_{\mathbf{D} \in \mathcal{D}(\mathcal{G}^*(\mathbf{Y}^*, \mathbf{A}^*))} \prod_{V \in \mathbf{D} \cup \mathbf{H}_{\mathbf{D}}} p(V \mid \pas_{\mathcal{G}(\mathbf{H} \cup \mathbf{V})}(V))  \\
        &= \prod_{\mathbf{D} \in \mathcal{D}(\mathcal{G}^*(\mathbf{Y}^*, \mathbf{A}^*))} \left(\sum_{\mathbf{H}_{\mathbf{D}}} \prod_{V \in \mathbf{D} \cup \mathbf{H}_{\mathbf{D}}} p(V \mid \pas_{\mathcal{G}(\mathbf{H} \cup \mathbf{V}} (V)) \right)
    \end{align*}

    The first equality follows from applying g-formula to a DAG and then marginalizing hidden variables $\mathbf{H}$. The second equality follows from Corollary~\ref{cor:c} since these hidden variables do not point into any $\mathbf{D} \in \mathcal{D}(\mathcal{G}^*(\mathbf{Y}^*, \mathbf{A}^*))$. This means that $V \in \mathbf{H}^* \cup \mathbf{Y}^*$ do not have parents from $\mathbf{H} \setminus \mathbf{H}^*$. The third equality follows from Corollary~\ref{cor:a} which allows us to factor the $V \in \mathbf{H}^* \cup \mathbf{Y}^*$ by districts. The fourth equality follows from Corollary~\ref{cor:b}, which allows us to partition the $\mathbf{H}^*$  by their districts under the sum.

    For any given $\mathbf{D}$, 
    \begin{align*}
        &\sum_{\mathbf{H}_{\mathbf{D}}} \prod_{V \in \mathbf{D} \cup \mathbf{H}_{\mathbf{D}}} p(V \mid \pas_{\mathcal{G}(\mathbf{H} \cup \mathbf{V})} (v)) \\
        &= \sum_{\mathbf{H}_{\mathbf{D}}} \prod_{V \in \mathbf{D} \cup \mathbf{H}_{\mathbf{D}} } p(V \mid \pas_{\mathcal{G}(\mathbf{H} \cup \mathbf{V})} (V)) \times\underbrace{\sum_{\mathbf{H} \setminus \mathbf{H}_{\mathbf{D}}} \prod_{v \in \mathbf{H} \setminus \mathbf{H}_{\mathbf{D}}} p(V \mid \pas_{\mathcal{G}(\mathbf{H} \cup \mathbf{V})} (V))}_{=1} \\
        &=\sum_{\mathbf{H}} \prod_{V \in \mathbf{D} \cup \mathbf{H}} p(V \mid \pas_{\mathcal{G}(\mathbf{H} \cup \mathbf{V})} (V)) \\
        &= \sum_{\mathbf{H}} \phi_{\mathbf{V} \setminus \mathbf{D}}(p(\mathbf{H} \cup \mathbf{V}); \mathcal{G}(\mathbf{H} \cup \mathbf{V}))
    \end{align*}

    The first equality 
    follows from Corollaries \ref{cor:b}, and \ref{cor:c}, the definition of $\mathbf{H}_{\mathbf{D}}$, and
    the fact that $\pas_{\mathcal{G}(\mathbf{H} \cup \mathbf{V}) }( \mathbf{D} \cup \mathbf{H}_{\mathbf{D}}) \cap (\mathbf{H} \setminus \mathbf{H}_{\mathbf{D}}) = \emptyset$.

	To see that this last fact is true, let $K \in \mathbf{H} \setminus \mathbf{H}_{\mathbf{D}}$. We show by contradiction that this $K$ cannot also be in $\pas_{\mathcal{G}(\mathbf{H} \cup \mathbf{V}) }( \mathbf{D} \cup \mathbf{H}_{\mathbf{D}})$    
    If $K$ is a parent of $\mathbf{D}$, then $K \in \mathbf{H}_{\mathbf{D}}$ since $K$ is a hidden variable with a direct path to a district. This is a contradiction.
    If $K$ is a parent of $\mathbf{H}_{\mathbf{D}}$, then there exists a path from $K$ to some element $H \in \mathbf{H}_{\mathbf{D}}$, and from $H$ to $\mathbf{D}$. In particular, all intermediate elements of this path are hidden. Therefore there is a directed path of hidden variables from $K$ to $\mathbf{D}$, which implies that $K \in \mathbf{H}_{\mathbf{D}}$. This is also a contradiction.

    Lemma 55 of \citet{richardsonNestedMarkovProperties2017} showed that fixing a set $\mathbf{A}$ and marginalizing a set $\mathbf{H}$ in a kernel $q_{\mathbf{V}}(\mathbf{V} \cup \mathbf{H} | \mathbf{W})$ commutes, provided this kernel is nested Markov relative to a CADMG ${\cal G}(\mathbf{V} \cup \mathbf{H}, \mathbf{W})$, and $\mathbf{A}$ is a fixable set in the latent projection ${\cal G}(\mathbf{V}, \mathbf{W})$.
    It follows that
    \begin{equation}
        \sum_{\mathbf{H}} \phi_{\mathbf{V} \setminus \mathbf{D}} (p(\mathbf{H} \cup \mathbf{V}); \mathcal{G}(\mathbf{H} \cup \mathbf{V})) = \phi_{\mathbf{V} \setminus \mathbf{D}} (p(\mathbf{V}); \mathcal{G}(\mathbf{V})) \label{eqn:lem55}
    \end{equation}
    and
    \begin{align*}
        &p(\mathbf{Y}^* \mid \doo_{\mathcal{G}(\mathbf{H} \cup \mathbf{V}) } (\mathbf{A}^*)) 
        = \prod_{\mathbf{D} \in \mathcal{D}(\mathcal{G}^*(\mathbf{Y}^*, \mathbf{A}^*)) } \phi_{\mathbf{V} \setminus \mathbf{D}} (p(\mathbf{V}); \mathcal{G}(\mathbf{V})).
    \end{align*}
    The conclusion follows since
    \begin{align*}
        &p(\mathbf{Y} \mid \doo_{\mathcal{G}(\mathbf{H} \cup \mathbf{V})} (\mathbf{A})) 
        = \sum_{\mathbf{Y}^* \setminus \mathbf{Y}} p(\mathbf{Y}^* \mid \doo_{\mathcal{G}(\mathbf{H} \cup \mathbf{V})} (\mathbf{A}^*)).
    \end{align*}

\end{prf}

\begin{lem}
	Fix a causal model associated with a DAG ${\cal G}(\mathbf{V} \cup \mathbf{H})$ defined on all one-step-ahead counterfactuals of the form $V(\mathbf{b}_V)$, where $\mathbf{b}_V$ are elements in
	${\mathfrak X}_{\pa_{{\cal G}(\mathbf{V} \cup \mathbf{H})}(V)}$ and $V \in \mathbf{V} \cup \mathbf{H}$.
	
	Consider the marginal SWIG ${\cal G}(\{ \mathbf{S} \cup \mathbf{H} \}(\mathbf{a}))$ for any $\mathbf{A} \subseteq \mathbf{V}$, $\mathbf{S} \equiv \mathbf{V} \setminus \mathbf{A}$.
	Then the set of counterfactuals $V(\mathbf{b}_V)$, where $\mathbf{b}_V$ are elements in
	${\mathfrak X}_{\pa_{{\cal G}(\{ \mathbf{S} \cup \mathbf{H} \}(\mathbf{a}))}}$ consistent with $\mathbf{a}$, and $V \in \mathbf{S} \cup \mathbf{H}$ forms a causal model associated with a (conditional) DAG 
	${\cal G}(\{ \mathbf{S} \cup \mathbf{H} \}(\mathbf{a}))$.
	
	Moreover, any counterfactual random variable $\mathbf{T}(\mathbf{c})$, where $\mathbf{T} \subseteq \mathbf{S} \cup \mathbf{H}$, and $\mathbf{a} \subseteq \mathbf{c}$ is equal in the two causal models.
\label{lem:sub-causal}
\end{lem}
\begin{prf}
This follows by standard structural equation model semantics of causal models, coupled with definition of interventions via structural equation replacement.
\end{prf}

\begin{lem}
	Fix a causal model associated with a DAG ${\cal G}(\mathbf{V} \cup \mathbf{H})$ defined on all one-step-ahead counterfactuals of the form $V(\mathbf{b}_V)$, where $\mathbf{b}_V$ are elements in
	${\mathfrak X}_{\pa_{{\cal G}(\mathbf{V} \cup \mathbf{H})}(V)}$ and $V \in \mathbf{V} \cup \mathbf{H}$.
	
	Consider a subset $\mathbf{A} \subseteq \mathbf{V} \cup \mathbf{H}$ ancestral in ${\cal G}(\mathbf{V} \cup \mathbf{H})$.
	Then the set of counterfactuals $V(\mathbf{b}_V)$, where $\mathbf{b}_V$ are elements in
	${\mathfrak X}_{\pa_{{\cal G}(\mathbf{A})}}$, and $V \in \mathbf{A}$ forms a causal model associated with ${\cal G}(\mathbf{A})$.

	Moreover, any counterfactual random variable $\mathbf{T}(\mathbf{c})$, where $\mathbf{T},\mathbf{C} \subseteq \mathbf{A}$ is equal in the two causal models.
\label{lem:an-causal}
\end{lem}
\begin{prf}
This follows by standard structural equation model semantics of causal models, coupled with definition of ancestral sets.
\end{prf}

\begin{lema}{\ref{lem:gid}}
    Fix a hidden variable causal model represented by an ADMG ${\cal G}(\mathbf{V})$, and a set of interventional distributions $\mathbb{Z} \equiv \{p(\mathbf{V}_i (\mathbf{b}_i))\}_{i=1}^k$  where $\mathbf{V}_i = \mathbf{V} \setminus \mathbf{B}_i$.   Then, $p(\mathbf{Y}(\mathbf{a}))$ is identified if and only if for each $\mathbf{D} \in {\cal D}({\cal G}(\mathbf{Y}^* (\mathbf{a})))$ where $\mathbf{Y}^* \equiv \an_{\mathcal{G}(\mathbf{V}(\mathbf{a}))}(\mathbf{Y}) \setminus \mathbf{a}$, there exists at least one $i_{\mathbf{D}} \in \{1, \ldots, k\}$ such that $p(\mathbf{V}_{i_{\mathbf{D}}}(\mathbf{b}_{i_{\mathbf{D}}})) \in \mathbb{Z}$ and $\mathbf{D} \in {\cal I}({\cal G} (\mathbf{V}_{i_{\mathbf{D}}} (\mathbf{b}_{i_{\mathbf{D}}})))$.
    Moreover, if $p(\mathbf{Y}(\mathbf{a}))$ is identified, it is equal to:
    \begin{align*}
        \sum_{\mathbf{Y}^* \setminus \mathbf{Y}}
        \!\!
        \prod_{\mathbf{D} \in {\cal D}({\cal G}({\mathbf{Y}^*}(\mathbf{a})))}
        \!\!\!\!\!\!\!\!
        \!\!\!\!
        \!\!
        \phi_{\mathbf{V}_{i_{\mathbf{D}}} \setminus \mathbf{D}}(p(\mathbf{V}_{i_{\mathbf{D}}}(\mathbf{b}_{i_{\mathbf{D}}})); {\cal G}(\mathbf{V}_{i_{\mathbf{D}}}(\mathbf{b}_{i_{\mathbf{D}}}))) \vert_{\mathbf{A} = \mathbf{a}}.
    \end{align*}
\end{lema}

\begin{prf}	
	Under the given causal model,
	\begin{align*}
	p(\mathbf{Y}(\mathbf{a})) = \sum_{\mathbf{Y}^* \setminus \mathbf{Y}} \prod_{\mathbf{D} \in {\cal D}({\cal G}(\mathbf{Y}^*(\mathbf{a})))} p(\mathbf{D}(\pa^s_{{\cal G}}(\mathbf{D}))) \vert_{\mathbf{A}=\mathbf{a}}.
	\end{align*}

	Lemma \ref{lem:sub-causal} immediately implies, by Theorem \ref{thm:id}, that the ID algorithm is sound applied to the marginal SWIG ${\cal G}(\{ \mathbf{V} \setminus \mathbf{A} \}(\mathbf{a}))$ and any 
	corresponding interventional distribution $p(\mathbf{S}(\mathbf{b}))$, provided $\mathbf{S} \subseteq \mathbf{V} \setminus \mathbf{A}$, and $\mathbf{a} \subseteq \mathbf{b}$.

    Fix $\mathbf{D} \in {\cal D}({\cal G}({\mathbf{Y}^*(\mathbf{a})}))$, and let $p(\mathbf{V}_{i_{\mathbf{D}}} (\mathbf{b}_{i_{\mathbf{D}}}))$ be the distribution used in the algorithm to obtain
    $p(\mathbf{D}(\pa^s_{{\cal G}}(\mathbf{D})))$.
	Then
	\begin{align*}
	p(\mathbf{D}(\pa^s_{{\cal G}}(\mathbf{D}))) = \phi_{\mathbf{V}_{i_{\mathbf{D}}} \setminus \mathbf{D}}( p(\mathbf{V}_{i_{\mathbf{D}}}); {\cal G}(\mathbf{V}_{i_{\mathbf{D}}}(\mathbf{b}_{i_{\mathbf D}})) ) \vert_{\mathbf{A}={\mathbf a}},
	\end{align*}
	since this is simply the output of the ID algorithm applied to the query $p(\mathbf{D}(\pa^s_{{\cal G}}(\mathbf{D})))$ with a known distribution $p(\mathbf{V}_{i_{\mathbf{D}}}(\mathbf{b}_{i_{\mathbf{D}}}))$ and the corresponding latent projection graph ${\cal G}(\mathbf{b}_{i_{\mathbf{D}}}))$ as inputs.  The fact that $\mathbf{D} \in {\cal I}({\cal G}(\mathbf{V}_{i_{\mathbf D}}(\mathbf{b}_{i_{\mathbf{D}}})))$ implies $p(\mathbf{D}(\pa^s_{{\cal G}}(\mathbf{D})))$ is identifiable from $p(\mathbf{V}_{i_{\mathbf{D}}}(\mathbf{b}_{i_{\mathbf{D}}}))$, and Lemma \ref{lem:sub-causal} implies $p(\mathbf{D}(\pa^s_{{\cal G}}(\mathbf{D})))$ is invariant in the original causal model yielding the latent projection ${\cal G}(\mathbf{V})$, and the new causal model yielding the marginal SWIG ${\cal G}(\mathbf{V}_{i_{\mathbf D}}(\mathbf{b}_{i_{\mathbf{D}}}))$.
	
\end{prf}

\begin{lema}{\ref{lem:aid}}
        Fix a hidden variable causal model represented by an ADMG ${\cal G}(\mathbf{V})$, and a set of interventional distributions
        $\mathbb{Z} \equiv \{p(\mathbf{S}_{i} (\mathbf{b}_{i} ))\}_{i=1}^k$,
        such that each $\mathbf{S}_i(\mathbf{b}_i)$ is ancestral in ${\cal G}(\mathbf{V}(\mathbf{a}))$.  Then $p(\mathbf{Y}(\mathbf{a}))$ is identified if and only if for each $\mathbf{D} \in {\cal D}({\cal G}(\mathbf{Y}^*(\mathbf{a})))$ where $\mathbf{Y}^* \equiv \an_{{\cal G}(\mathbf{V}(\mathbf{a}))}(\mathbf{Y}) \setminus \mathbf{a}$, there exists at least one $i_{\mathbf{D}} \in \{ 1, \ldots, k \}$ such that $p(\mathbf{S}_{i_{\mathbf{D}}}(\mathbf{b}_{i_{\mathbf{D}}})) \in {\mathbb Z}$ and
        $\mathbf{D} \in {\cal I}({\cal G}({\mathbf S}_{i_{\mathbf{D}}}(\mathbf{b}_{i_{\mathbf{D}}})))$.
        Moreover, if $p(\mathbf{Y}(\mathbf{a}))$ is identified, it is equal to:
        \begin{align*}
            \sum_{\mathbf{Y}^* \setminus \mathbf{Y}} \prod_{\mathbf{D} \in {\cal D}({\cal G}({\mathbf{Y}^*}(\mathbf{a})))}
            \!\!\!\!\!\!\!\!
            \!\!\!\!\!
            \phi_{\mathbf{S}_{i_{\mathbf{D}}} \setminus \mathbf{D}}(p(\mathbf{S}_{i_{\mathbf{D}}}(\mathbf{b}_{i_{\mathbf{D}}})); {\cal G}(\mathbf{S}_{i_{\mathbf{D}}}(\mathbf{b}_{i_{\mathbf{D}}}))) \vert_{\mathbf{A} = \mathbf{a}}.
        \end{align*}
\end{lema}

\begin{prf}
	Under the given causal model,
	\begin{align*}
	p(\mathbf{Y}(\mathbf{a})) = \sum_{\mathbf{Y}^* \setminus \mathbf{Y}} \prod_{\mathbf{D} \in {\cal D}({\cal G}(\mathbf{Y}^*(\mathbf{a})))} p(\mathbf{D}(\pa^s_{{\cal G}}(\mathbf{D}))) \vert_{\mathbf{A}=\mathbf{a}}.
	\end{align*}

	Fix $\mathbf{D} \in {\cal D}({\cal G}(\mathbf{Y}^*(\mathbf{a})))$, and the corresponding $p(\mathbf{S}_{i_{\mathbf{D}}}(\mathbf{b}_{i_{\mathbf{D}}})) \in \mathbb{Z}$.
	Fix any ${\cal G}(\mathbf{V} \cup \mathbf{H})$ representing a causal model, of which ${\cal G}(\mathbf{V})$ is a latent projection.

	Let $\mathbf{S}^* \equiv \an_{{\cal G}(\mathbf{V} \cup \mathbf{H})}(\mathbf{S}_{i_{\mathbf{D}}})$.  By definition, the set $\mathbf{S}^*$ is ancestral in ${\cal G}(\mathbf{V} \cup \mathbf{H})$.  By Lemma \ref{lem:an-causal},
	the DAG ${\cal G}(\mathbf{S}^*)$ forms a causal model with respect to the appropriate set of counterfactuals.  By Lemma \ref{lem:sub-causal},
	the marginal SWIG ${\cal G}(\{ \mathbf{S}^* \setminus \mathbf{B}_{i_{\mathbf D}} \}(\mathbf{b}_{i_{\mathbf D}}))$ forms a causal model with respect to the appropriate set of counterfactuals.
	
	This implies the ID algorithm is sound applied to the marginal SWIG ${\cal G}(\mathbf{S}_{i_{\mathbf D}}(\mathbf{b}_{i_{\mathbf D}}))$ obtained as a latent projection of ${\cal G}(\{ \mathbf{S}^* \setminus \mathbf{B}_{i_{\mathbf D}} \}(\mathbf{b}_{i_{\mathbf D}}))$, and any corresponding interventional distribution $p(\mathbf{S}_{i_{\mathbf D}}(\mathbf{b}_{i_{\mathbf D}}))$.  This implies that in the causal model associated with the marginal
	SWIG ${\cal G}(\mathbf{S}_{i_{\mathbf D}}(\mathbf{b}_{i_{\mathbf D}}))$,
	\begin{align*}
	p(\mathbf{D}(\pa^s_{{\cal G}}(\mathbf{D}))) = \phi_{\mathbf{V}_{i_{\mathbf{D}}} \setminus \mathbf{D}}( p(\mathbf{S}_{i_{\mathbf{D}}}); {\cal G}(\mathbf{S}_{i_{\mathbf{D}}}(\mathbf{b}_{i_{\mathbf D}})) ) \vert_{\mathbf{A}={\mathbf a}}.
	\end{align*}
	Lemma \ref{lem:an-causal} implies $p(\mathbf{D}(\pa^s_{{\cal G}}(\mathbf{D})))$ is invariant in the original causal model yielding the latent projection ${\cal G}(\mathbf{V})$, and the new causal model yielding the marginal SWIG ${\cal G}(\mathbf{S}_{i_{\mathbf D}}(\mathbf{b}_{i_{\mathbf{D}}}))$.  This establishes our conclusion.

\end{prf}
\begin{lema}{\ref{lem:mid}}
    Given a hidden variable causal model represented by an ADMG ${\cal G}(\mathbf{V})$, and a set of interventional distributions
    $\mathbb{Z} \equiv \{p(\mathbf{S}_{i} (\mathbf{b}_{i} ))\}_{i=1}^k$,
    then $p(\mathbf{Y}(\mathbf{a}))$ is identified from $\mathbb{Z}$ if there exists $\mathbf{Y}' \subseteq {\cal P}(\mathbf{Y}^* \setminus \mathbf{Y}) \cup \mathbf{Y}$,\footnote{${\cal P}(\mathbf{S})$ for any set $\mathbf{S}$ is the power set of $\mathbf{S}$.} such that for each $\mathbf{D} \in {\cal D}({\cal G}(\mathbf{Y}'(\mathbf{a})))$, we can find at least one $i_{\mathbf{D}}$ such that $p(\mathbf{S}_{i_{\mathbf{D}}}(\mathbf{b}_{i_{\mathbf{D}}})) \in {\mathbb Z}$,
    $\pas_{\mathcal{G}(\mathbf{S}_{i_{\mathbf{D}}}(\mathbf{b}_{i_{\mathbf{D}}}))} (\mathbf{D}) = \pas_{\mathcal{G}(\mathbf{Y}')}(\mathbf{D})$, and $\mathbf{D} \in {\cal I}({\cal G}(\mathbf{S}_{i_{\mathbf{D}}} (\mathbf{b}_{i_{\mathbf{D}}})))$.
    Moreover, if $p(\mathbf{Y}(\mathbf{a}))$ is identified, the identifying formula is:
    \begin{align*}
        \sum_{\mathbf{Y}' \setminus \mathbf{Y}} \prod_{\mathbf{D} \in {\cal D}({\cal G}({\mathbf{Y}'}(\mathbf{a})))}
        \!\!\!\!\!\!\!\!
        \!\!\!\!
        \phi_{\mathbf{S}_{i_{\mathbf{D}}} \setminus \mathbf{D}}(p(\mathbf{S}_{i_{\mathbf{D}}}(\mathbf{b}_{i_{\mathbf{D}}})); {\cal G}(\mathbf{S}_{i_{\mathbf{D}}}(\mathbf{b}_{i_{\mathbf{D}}}))) \vert_{\mathbf{A} = \mathbf{a}}.
    \end{align*}

\end{lema}
\begin{prf}
	Under the given causal model,
	\begin{align*}
	p(\mathbf{Y}(\mathbf{a})) = \sum_{\mathbf{Y}' \setminus \mathbf{Y}} \prod_{\mathbf{D} \in {\cal D}({\cal G}(\mathbf{Y}'(\mathbf{a})))} p(\mathbf{D}(\pa^s_{{\cal G}}(\mathbf{D}))) \vert_{\mathbf{A}=\mathbf{a}}.
	\end{align*}
	Note here that unlike previous proofs, the SWIG ${\cal G}(\mathbf{Y}'(\mathbf{a}))$ is potentially a latent projection of the SWIG ${\cal G}(\mathbf{Y}^*(\mathbf{a}))$.  Nevertheless, since
	in any underlying causal model associated with a hidden variable DAG ${\cal G}(\mathbf{V} \cup \mathbf{H})$, the distribution $p(\mathbf{V}(\mathbf{a}) \cup \mathbf{H}(\mathbf{a})$ is Markov relative to ${\cal G}(\mathbf{V}(\mathbf{a}) \cup \mathbf{H}(\mathbf{a}))$, implies $p(\mathbf{Y}'(\mathbf{a}))$ is nested Markov relative to ${\cal G}(\mathbf{Y}'(\mathbf{a}))$, which in turn implies the district factorization above.

	Fix $\mathbf{D} \in {\cal D}({\cal G}(\mathbf{Y}'(\mathbf{a})))$, and the corresponding $p(\mathbf{S}_{i_{\mathbf{D}}}(\mathbf{b}_{i_{\mathbf{D}}})) \in \mathbb{Z}$.
	Fix any ${\cal G}(\mathbf{V} \cup \mathbf{H})$ representing a causal model, of which ${\cal G}(\mathbf{V})$ is a latent projection.

	Let $\mathbf{S}^* \equiv \an_{{\cal G}(\mathbf{V} \cup \mathbf{H})}(\mathbf{S}_{i_{\mathbf{D}}})$.  By definition, the set $\mathbf{S}^*$ is ancestral in ${\cal G}(\mathbf{V} \cup \mathbf{H})$.  By Lemma \ref{lem:an-causal},
	the DAG ${\cal G}(\mathbf{S}^*)$ forms a causal model with respect to the appropriate set of counterfactuals.  By Lemma \ref{lem:sub-causal},
	the marginal SWIG ${\cal G}(\{ \mathbf{S}^* \setminus \mathbf{B}_{i_{\mathbf D}} \}(\mathbf{b}_{i_{\mathbf D}}))$ forms a causal model with respect to the appropriate set of counterfactuals.
	
	This implies the ID algorithm is sound applied to the marginal SWIG ${\cal G}(\mathbf{S}_{i_{\mathbf D}}(\mathbf{b}_{i_{\mathbf D}}))$ obtained as a latent projection of ${\cal G}(\{ \mathbf{S}^* \setminus \mathbf{B}_{i_{\mathbf D}} \}(\mathbf{b}_{i_{\mathbf D}}))$, and any corresponding interventional distribution $p(\mathbf{S}_{i_{\mathbf D}}(\mathbf{b}_{i_{\mathbf D}}))$.
	This implies that in the causal model associated with the marginal
	SWIG ${\cal G}(\mathbf{S}_{i_{\mathbf D}}(\mathbf{b}_{i_{\mathbf D}}))$,
	\begin{align*}
	p(\mathbf{D}(\pa^s_{{\cal G}(\mathbf{Y}'(\mathbf{a}))}(\mathbf{D}))) = \phi_{\mathbf{V}_{i_{\mathbf{D}}} \setminus \mathbf{D}}( p(\mathbf{S}_{i_{\mathbf{D}}}); {\cal G}(\mathbf{S}_{i_{\mathbf{D}}}(\mathbf{b}_{i_{\mathbf D}})) ) \vert_{\mathbf{A}={\mathbf a}}.
	\end{align*}
	That $p(\mathbf{D}(\pa^s_{{\cal G}(\mathbf{Y}'(\mathbf{a}))}(\mathbf{D})))$ is invariant in the original causal model yielding the latent projection ${\cal G}(\mathbf{V})$, and the new causal model yielding the marginal SWIG ${\cal G}(\mathbf{S}_{i_{\mathbf D}}(\mathbf{b}_{i_{\mathbf{D}}}))$ follows from the assumption that $\pa^s_{{\cal G}(\mathbf{Y}'(\mathbf{a}))}(\mathbf{D})) = \pa^s_{{\cal G}(\mathbf{S}_{i_{\mathbf D}}(\mathbf{b}_{i_{\mathbf D}}))}(\mathbf{D})$.
	This establishes our conclusion.

\end{prf}

\begin{lema}{\ref{lem:commute}}
    If $A \subseteq \mathbf{V}$ is s-fixable in a CADMG ${\cal G}(\mathbf{V},\mathbf{W})$ where $\mathbf{C}\subseteq\mathbf{V}$ is conditioned, associated with a kernel $q_{\mathbf V}(\mathbf{V} \setminus \mathbf{C} \mid \mathbf{W} = \mathbf{w}, \mathbf{C} = \mathbf{c})$, then
    \begin{align*}
        q_{\mathbf{V}\setminus\{A\}}(\mathbf{V}\setminus(\mathbf{C}\cup\{A\})|\mathbf{W}\cup\{A\}, \mathbf{C}=\mathbf{c}) =\\
        \phi^{\mathbf{C}}_A(q_{\mathbf{V}}(\mathbf{V}\setminus\mathbf{C} | \mathbf{W},\mathbf{C}=\mathbf{c}), {\cal G}(\mathbf{V},\mathbf{W}))\text{, with}\\
        q_{\mathbf{V}\setminus\{A\}}(\mathbf{V}\setminus\{A\}|\mathbf{W}\cup\{A\}) \equiv \phi_A(q_{\mathbf{V}}(\mathbf{V} | \mathbf{W}), {\cal G}(\mathbf{V},\mathbf{W})).
    \end{align*}
\end{lema}
\begin{prf}
    Consider first conditioning on $\mathbf{C}= \mathbf{c}$ in $q_{\mathbf{V}}(\mathbf{V} \mid \mathbf{W})$. By the definition of conditioning in a kernel, we obtain 
    \[q_{\mathbf{V}}(\mathbf{V} \setminus \mathbf{C} \mid \mathbf{W}, \mathbf{C} = \mathbf{c}) = \frac{q_{\mathbf{V}}(\mathbf{V} \mid \mathbf{W})}{\sum_{\mathbf{V} \setminus \mathbf{C} }q_{\mathbf{V}}(\mathbf{V} \mid \mathbf{W})}\Big\vert_{\mathbf{C} = \mathbf{c}}\]

    The subsequent s-fixing on $A$ results in
    \begin{align*}
        &\frac{q_{\mathbf{V}}(\mathbf{V} \setminus \mathbf{C} \mid \mathbf{W}, \mathbf{C} =\mathbf{c})}{q_{\mathbf{V}} (A \mid \mathbf{W}, \nd_{\cal G}(A) \setminus (\mathbf{C}\cup\mathbf{W}), \mathbf{C}= \mathbf{c})} \\
        &= q_{\mathbf{V}}(\mathbf{V} \setminus (\nd_{\cal G}(A) \dot{\cup} \{A\}) \mid A, \nd_{\cal G}(A) \setminus (\mathbf{C}\cup\mathbf{W}), \mathbf{C} = \mathbf{c},\mathbf{W})  \\
        &\quad\times  q_{\mathbf{V}}(\nd_{\cal G}(A) \setminus (\mathbf{C}\cup\mathbf{W}) \mid \mathbf{W}, \mathbf{C} = \mathbf{c})
    \end{align*}

    where $\mathbf{C} \subseteq \nd_{\cal G}(A)$ by the definition of s-fixing, and where equation 10 in \citet{richardsonNestedMarkovProperties2017} was applied.

    Conversely, consider first fixing $A$ in $q_{\mathbf{V}} (\mathbf{V} \mid \mathbf{W})$. This is defined as
    \begin{align*}
        &\frac{q_{\mathbf{V}}(\mathbf{V} \mid \mathbf{W})}{q_{\mathbf{V}} (A \mid \nd_{\cal G}(A)\setminus\mathbf{W}, \mathbf{W})} \\
        &= q_{\mathbf{V}}(\mathbf{V} \setminus (\nd_{\cal G}(A) \cup \{A\}) \mid A, \nd_{\cal G}(A))\setminus\mathbf{W}, \mathbf{W})\\
        & \quad \times q_{\mathbf{V}}(\nd(A))\setminus\mathbf{W} \mid \mathbf{W})
    \end{align*}

    by equation 10 of \citet{richardsonNestedMarkovProperties2017}, where $\mathbf{C} \subseteq \nd_{\cal G}(A)$.

    Conditioning on $C$ and setting $\mathbf{C} = \mathbf{c}$ in this distribution gives 
    \begin{align*}
        &q_{\mathbf{V}}(\mathbf{V} \setminus (\nd_{\cal G}(A) \cup \{A\}) \mid A, \nd_{\cal G}(A) \setminus (\mathbf{C}\cup\mathbf{W}), \mathbf{C} = \mathbf{c}, \mathbf{W})\\
        & \quad \times q_{\mathbf{V}}(\nd(A) \setminus (\mathbf{C}\cup\mathbf{W}) \mid \mathbf{W}, \mathbf{C} = \mathbf{c})
    \end{align*}
    by applying the definition of conditioning, which demonstrates the required commutativity property.
    
\end{prf}

\begin{lema}{\ref{lem:joint-from-conditional}}
 Given a SWIG ${\cal G}(\mathbf{S}(\mathbf{b}))$, and the corresponding interventional distribution $p(\mathbf{S}(\mathbf{b}))$,
 let $\mathbf{Y}^* = (\an_{{\cal G}(\mathbf{S}(\mathbf{b}))}(\mathbf{Y}) \setminus \mathbf{a})$. Let $\mathbf{Y} \subseteq \mathbf{S}$,
 and $\mathbf{A} \subseteq \mathbf{S} \cup \mathbf{B}$.
 $p(\mathbf{Y}(\mathbf{a}))$ is identified from $p(\{\mathbf{S} \setminus \mathbf{C}\}(\mathbf{b}) | \mathbf{C}(\mathbf{b}) = \mathbf{c})$
 if $\de_{{\cal G}({\bf S}({\bf b}))}(\mathbf{Y}^*) \cap \mathbf{C} = \emptyset$,
 $\mathbf{c}_{\pa_{\cal G}(\mathbf{Y}^*) \cap \mathbf{A}}$ is consistent with $\mathbf{a}$,
 $\mathbf{b}_{\pa_{\cal G}(\mathbf{Y}^*) \cap \mathbf{A}}$ is consistent with $\mathbf{a}$,
 and for each district $\mathbf{D} \in {\cal D}({\cal G}(\mathbf{Y}^*(\mathbf{a})))$,
 there exists a set ${\bf Z}_{\bf D} \in {\bf S} \setminus {\bf C}$, such that
 ${\bf D} \subseteq {\bf Z}_{\bf D}$,
${\bf Z}_{\bf D}$ is s-fixable in ${\cal G}(\mathbf{S}(\mathbf{b}))$ (given a conditioned ${\bf C}$) by a sequence
 $\sigma_{{\bf Z}_{\bf D}}$ that fixes $\bar{\bf D} \subseteq {\bf Z}_{\bf D}$ last, where $\bar{\bf D}$ is a district in $\phi^{\bf C}_{\sigma_{{\bf Z}_{\bf D} \setminus \bar{\bf D}}}({\cal G}({\bf S}({\bf b})))$, and
 ${\bf D}$ is reachable in $\phi^{\bf C}_{\sigma_{{\bf Z}_{\bf D} \setminus \bar{\bf D}}}({\cal G}({\bf S}({\bf b})))$.

If $p(\mathbf{Y}(\mathbf{a}))$ is identified, it is equal to
\begin{align*}
    \sum_{\mathbf{Y}^* \setminus \mathbf{Y}} \prod_{\mathbf{D} \in {\cal D}({\cal G}(\mathbf{Y}^*({\mathbf a})))} q_{\bf D}({\bf D} | \pa^s_{{\cal G}({\bf S}({\bf b}))}({\bf D})) \vert_{{\bf A}={\bf a}},
\end{align*}
where for each ${\bf D} \in {\cal D}({\cal G}(\mathbf{Y}^*({\mathbf a})))$,
{\small
\begin{align*}
q_{\bf D}({\bf D} | \pa^s_{{\cal G}({\bf S}({\bf b}))}({\bf D})) \equiv \phi_{\bar{\bf D} \setminus {\bf D}}(q_{\bar{\bf D}}(\bar{\bf D} | \pa^s_{{\cal G}}(\bar{\bf D})); \phi_{{\bf S} \setminus \bar{\bf D}}({\cal G}({\bf S}({\bf b}))))\\
q_{\bar{\bf D}}(\bar{\bf D} | \pa^s_{{\cal G}}(\bar{\bf D})) \equiv \prod_{D \in \bar{\bf D}} q_{{\bf S} \setminus ({\bf Z}_{\bf D} \setminus \bar{\bf D})}(D | \mb^*(D),
{\bf Z}_{\bf D} \setminus \bar{\bf D})\\
                    q_{{\bf S} \setminus ({\bf Z}_{\bf D} \!\setminus\! \bar{\bf D})}({\bf S} \!\setminus\! ({\bf Z}_{\bf D} \!\setminus\! \bar{\bf D}) | {\bf Z}_{\bf D} \!\setminus\! \bar{\bf D}) \equiv \\
                    	\phi^{\bf C}_{{\bf Z}_{\bf D} \!\setminus\! \bar{\bf D}}(p(\{ {\bf S} \setminus {\bf C} \}({\bf b} | {\bf C}({\bf b}) = {\bf c}); {\cal G}({\bf S}({\bf b})))),
\end{align*}
}
with $\mb^*(D)$ defined as $\mb_{\phi_{{\bf Z}_{\bf D} \setminus \bar{\bf D}}({\cal G}({\bf S}({\bf b})))}(D)$ intersected with elements in $\bar{\bf D}$ earlier than $D$ in any reverse topological order in $\phi_{{\bf Z}_{\bf D} \setminus \bar{\bf D}}({\cal G}({\bf S}({\bf b})))$.
\end{lema}

\begin{prf}
Since s-fixability implies fixability, under the given causal model, we have
\begin{align*}
    \sum_{\mathbf{Y}^* \setminus \mathbf{Y}} \prod_{\mathbf{D} \in {\cal D}({\cal G}(\mathbf{Y}^*({\mathbf a})))}\phi_{\sigma_{\mathbf{V} \setminus {D}}} (p(\mathbf{S}(\mathbf{b})), {\cal G}(\mathbf{S}(\mathbf{b}))) \vert_{\mathbf{A} = \mathbf{a}},
\end{align*}
for any valid sequence $\sigma_{\mathbf{V} \setminus \mathbf{D}}$ for every $\mathbf{D}$.

The kernel obtained by s-fixing ${\bf Z}_\mathbf{D} \setminus \bar{\bf D}$ is equal to the kernel we obtain by fixing had there been no selection bias, by inductive application of Lemma \ref{lem:commute}, and assumption of s-fixability of ${\bf Z}_\mathbf{D}$ by sequence $\sigma$ (and thus of  ${\bf Z}_\mathbf{D} \setminus \bar{\bf D}$, since elements in $\bar{\bf D}$ are fixed last by $\sigma$).

Since $p({\bf S}({\bf b}))$ is nested Markov with respect to ${\cal G}({\bf S}({\bf b}))$, and since $\bar{\bf D}$ is s-fixable once we s-fix ${\bf Z}_{\bf D} \setminus {\bf D}$,
the terms in $\prod_{D \in \bar{\bf D}} q_{{\bf S} \setminus ({\bf Z}_{\bf D} \setminus \bar{\bf D})}(D | \mb^*(D),{\bf Z}_{\bf D} \setminus \bar{\bf D})$ correspond, via chain rule, to the kernel
$q_{\bar{\bf D}}(\bar{\bf D} | \pa^s_{{\cal G}}(\bar{\bf D})) \equiv \phi_{{\bf S} \setminus \bar{\bf D}}(p({\bf S}({\bf b})); {\cal G}({\bf S}({\bf b})))$.  Specifically, this is established by induction by noting that
at every stage only childless unfixed elements in $\bar{\bf D}$ may be fixed, using the nested Markov property to note fixing every element in $\bar{\bf D}$ is equivalent to marginalization, and using chain rule.  That each such element is \emph{s-fixable} rather than just fixable implies the resulting object is independent of all variables in ${\bf C}$ and thus their product is equal to $\phi_{{\bf S} \setminus \bar{\bf D}}(p({\bf S}({\bf b})); {\cal G}({\bf S}({\bf b})))$ (note the usual fixing operator, rather than the s-fixing operator).

Since $p({\bf S}({\bf b}))$ is nested Markov with respect to ${\cal G}({\bf S}({\bf b}))$, this kernel is nested Markov relative to $\phi_{{\bf S} \setminus \bar{\bf D}}({\cal G}({\bf S}({\bf b})))$, which means the last step of the algorithm, namely $q_{\bf D}({\bf D} | \pa^s_{{\cal G}({\bf S}({\bf b}))}({\bf D})) \equiv \phi_{\bar{\bf D} \setminus {\bf D}}(q_{\bar{\bf D}}(\bar{\bf D} | \pa^s_{{\cal G}}(\bar{\bf D}));
\phi_{{\bf S} \setminus \bar{\bf D}}({\cal G}({\bf S}({\bf b}))))$, is sound since the ID algorithm expressed via the fixing operator is sound.  This has been shown in \cite{richardson17nested}.

\end{prf}

\begin{lema}{\ref{lem:eid_statement}}
    Fix a hidden variable causal model represented by an ADMG ${\cal G}(\mathbf{V})$, and a chain rule closed set of distributions ${\mathbb Z} = \{ p(\mathbf{S}_i(\mathbf{b}_i) | \mathbf{C}_i(\mathbf{b}_i)) \}_{i=1}^k$ (with some possibly available only at a level ${\bf c}_i$).
    Then $p(\mathbf{Y}(\mathbf{a}))$ is identified from $\mathbb{Z}$ if
    there exists $\mathbf{Y}' \subseteq {\cal P}(\mathbf{Y}^* \setminus \mathbf{Y}) \cup \mathbf{Y}$, such that for
    for each $\mathbf{D} \in {\cal D}({\cal G}(\mathbf{Y}'(\mathbf{a})))$, we can find at least one $i_{\mathbf{D}}$ such that
    $\pas_{\mathcal{G}(\{ \mathbf{S}_{i_{\mathbf{D}}} \cup \mathbf{C}_{i_{\mathbf{D}}} \}(\mathbf{b}_{i_{\mathbf{D}}}) )} (\mathbf{D}) = \pas_{\mathcal{G}(\mathbf{Y}')}(\mathbf{D})$,
    and
    $p(\mathbf{D} \mid \doo(\pas_{\mathcal{G}(\mathbf{S}_{i_{\mathbf{D}}}(\mathbf{b}_{i_{\mathbf{D}}}) \cup \mathbf{C}_{i_{\mathbf{D}}}(\mathbf{b}_{i_{\mathbf{D}}}))} (\mathbf{D})))$ is identified from $p(\mathbf{S}_i(\mathbf{b}_i) | \mathbf{C}_i(\mathbf{b}_i))$ evaluated at ${\bf c}_i$ consistent with ${\bf a}$ using Lemma \ref{lem:joint-from-conditional}.
    Moreover, if $p(\mathbf{Y}(\mathbf{a}))$ is identified, it is equal to:
    \begin{align*}
        \sum_{\mathbf{Y}' \setminus \mathbf{Y}} \prod_{\mathbf{D} \in {\cal D}({\cal G}({\mathbf{Y}'}(\mathbf{a})))}
        \!\!\!\!\!\!\!\!
        p(\mathbf{D} (\pas_{\mathcal{G}(\mathbf{S}_{i_{\mathbf{D}}}(\mathbf{b}_{i_{\mathbf{D}}}) \cup \mathbf{C}_{i_{\mathbf{D}}}(\mathbf{b}_{i_{\mathbf{D}}}))} (\mathbf{D}))) \vert_{\mathbf{A}=\mathbf{a}} \\
        =
        \sum_{\mathbf{Y}' \setminus \mathbf{Y}} \prod_{\mathbf{D} \in {\cal D}({\cal G}({\mathbf{Y}'}(\mathbf{a})))}
        q_{\mathbf{D}}(\mathbf{D} | \pas_{{\cal G}_{\mathbf{Y}'}}(\mathbf{D}) ) \vert_{\mathbf{A}=\mathbf{a}},
    \end{align*}
    where each $q_{\mathbf{D}}(\mathbf{D} | \pas_{{\cal G}_{\mathbf{Y}'}}(\mathbf{D}) )$ is obtained from applying Lemma \ref{lem:joint-from-conditional} to the appropriate element of $\mathbb{Z}$.
\end{lema}
\begin{prf}
	As before, under the given causal model,
	\begin{align*}
	p(\mathbf{Y}(\mathbf{a})) = \sum_{\mathbf{Y}' \setminus \mathbf{Y}} \prod_{\mathbf{D} \in {\cal D}({\cal G}(\mathbf{Y}'(\mathbf{a})))} p(\mathbf{D}(\pa^s_{{\cal G}}(\mathbf{D}))) \vert_{\mathbf{A}=\mathbf{a}}.
	\end{align*}

	Fix $\mathbf{D} \in {\cal D}({\cal G}(\mathbf{Y}'(\mathbf{a})))$, and the corresponding $p(\{ \mathbf{S}_{i_{\mathbf{D}}} \setminus \mathbf{C}_{{i_{\mathbf{D}}}} \}(\mathbf{b}_{i_{\mathbf{D}}}) | \mathbf{C}_{i_{\mathbf{D}}}(\mathbf{b}_{i_{\mathbf{D}}})) \in \mathbb{Z}$.  Fix any ${\cal G}(\mathbf{V} \cup \mathbf{H})$ representing a causal model, of which ${\cal G}(\mathbf{V})$ is a latent projection.

	Let $\mathbf{S}^* \equiv \an_{{\cal G}(\mathbf{V} \cup \mathbf{H})}(\mathbf{S}_{i_{\mathbf D}} \cup \mathbf{C}_{i_{\mathbf D}})$.
	By definition, the set $\mathbf{S}^*$ is ancestral in ${\cal G}(\mathbf{V} \cup \mathbf{H})$.  By Lemma \ref{lem:an-causal},
	the DAG ${\cal G}(\mathbf{S}^*)$ forms a causal model with respect to the appropriate set of counterfactuals.  By Lemma \ref{lem:sub-causal},
	the marginal SWIG ${\cal G}(\{ \mathbf{S}^* \setminus \mathbf{B}_{i_{\mathbf D}} \}(\mathbf{b}_{i_{\mathbf D}}))$ forms a causal model with respect to the appropriate set of counterfactuals.
	
	This implies that we can apply the results of Lemma \ref{lem:commute} to the marginal SWIG ${\cal G}(\{ \mathbf{S}_{i_{\mathbf D}} \cup \mathbf{C}_{i_{\mathbf D}} \}(\mathbf{b}_{i_{\mathbf D}}))$ obtained as a latent projection of ${\cal G}(\{ \mathbf{S}^* \setminus \mathbf{B}_{i_{\mathbf D}} \}(\mathbf{b}_{i_{\mathbf D}}))$, and any corresponding conditional interventional distribution
	$p(\mathbf{S}_{i_{\mathbf D}}(\mathbf{b}_{i_{\mathbf D}}) | \mathbf{C}_{i_{\mathbf D}}(\mathbf{b}_{i_{\mathbf D}}))$ to yield $p(\mathbf{D}(\pa^s_{{\cal G}(\mathbf{Y}'(\mathbf{a}))}(\mathbf{D})))$.
	That $p(\mathbf{D}(\pa^s_{{\cal G}(\mathbf{Y}'(\mathbf{a}))}(\mathbf{D})))$ is invariant in the original causal model yielding the latent projection ${\cal G}(\mathbf{V})$, and the new causal model yielding the marginal SWIG ${\cal G}(\mathbf{S}_{i_{\mathbf D}}(\mathbf{b}_{i_{\mathbf{D}}}))$ follows from the assumption that $\pa^s_{{\cal G}(\mathbf{Y}'(\mathbf{a}))}(\mathbf{D})) = \pa^s_{{\cal G}(\mathbf{S}_{i_{\mathbf D}}(\mathbf{b}_{i_{\mathbf D}}))}(\mathbf{D})$.
	This establishes our conclusion.
\end{prf}

\section{Proofs of Completeness}

\begin{lema}{\ref{lem:gid_fails}}
    If a causal query $p(\mathbf{Y}(\mathbf{a}))$ fails from $\mathbb{Z}$ using aID, and fails from $\bar{\mathbb{Z}}$ using gID, then this causal query is not identified from $\mathbb{Z}$.
\end{lema}

\begin{prf}
    Both aID and gID will fail on some intrinsic set $\mathbf{D} \in \mathcal{D}(\mathcal{G}_{\mathbf{Y}^*})$. For this intrinsic set $\mathbf{D}$, we can construct a thicket \cite{leeGeneralIdentifiabilityArbitrary2019} from the point where gID fails. On this thicket we can construct models $\mathcal{M}_1, \mathcal{M}_2$ with the property that they will agree on interventional distributions from $\bar{\mathbb{Z}}$ including $\mathbf{D}$ and $\pa(\mathbf{D})$, but $p^1(\mathbf{D} \mid \doo(\pa(\mathbf{D}))) \neq p^2(\mathbf{D} \mid \doo(\pa(\mathbf{D})))$. The construction will align on elements from $\mathbb{Z}$, as these are marginals of elements of $\bar{\mathbb{Z}}$. 
\end{prf}

\begin{lema}{\ref{lem:aid_fails}}
    If a causal query $p(\mathbf{Y}(\mathbf{a}))$ fails from $\mathbb{Z}$ using aID, but succeeds from $\bar{\mathbb Z}$ using gID, then this causal query is not identified from $\mathbb{Z}$. 
\end{lema}

\begin{prf}
    If identification fails for aID but succeeds for gID, then we can find at least one such required intrinsic set $\mathbf{D}$ with associated parents $\pa_{\cal G}(\mathbf{D})$ which exists in $\bar{\mathbb{Z}}$ but not in $\mathbb{Z}$. 

    For each $p(\mathbf{V}_i (\mathbf{b}_i)) \in \bar{\mathbb Z}$ and $p(\mathbf{S}_i (\mathbf{b}_i)) \in \mathbb{Z}$, the kernel $q_{\mathbf{V}_i}(\mathbf{D} \mid \mathbf{V}_i \setminus \mathbf{D})$ corresponding to intrinsic set and parents $\mathbf{D} \cup \pa_{\cal G}(\mathbf{D})$ is:
    \begin{enumerate}
        \item  absent in $p(\mathbf{V}_i (\mathbf{b}_i)) $ and $p(\mathbf{S}_i (\mathbf{b}_i)) $
        \item  present in $p(\mathbf{V}_i (\mathbf{b}_i)) $, but absent in $p(\mathbf{S}_i (\mathbf{b}_i)) $ because parts of $\mathbf{D} \cup \pa_{\cal G}(\mathbf{D})$ were marginalized
        \item  present in $p(\mathbf{V}_i (\mathbf{b}_i)) $, but absent in $p(\mathbf{S}_i (\mathbf{b}_i)) $ because $\mathbf{D}$ was not reachable
    \end{enumerate}
    where present means that there exist a valid fixing order for the kernel in the specified distribution, and absent means no such order exists.

    Note that by ancestrality of $\bar{\mathbb Z}$, $\pa_{{\cal G}_{\mathbf{Y}^*}  }(\mathbf{D})$ is equal to $\pa_{{\cal G}(\mathbf{V}_i(\mathbf{b}_i))}(\mathbf{D})$ if $\mathbf{D} \subseteq \mathbf{V}_i (\mathbf{b}_i)$ for all $i$, and similarly for if $\mathbf{V}_i(\mathbf{b}_i)$ is replaced with $\mathbf{S}_i (\mathbf{b}_i)$.

    We claim case 3 cannot happen. Consider the variables $\mathbf{D} \cup \pa_{\cal G}(\mathbf{D})$ which are present (but not reachable) in $p(\mathbf{S}_i (\mathbf{b}_i)) \in \mathbb{Z}$. $\mathbf{D}$ is reachable in some $p(\mathbf{V}_i (\mathbf{b}_i)) \in \bar{\mathbb Z}$, which is a distribution over possibly more variables. Further, by ancestrality of $\mathbb{Z}$, the set difference of $\mathbf{V}_i(\mathbf{b}_i) \setminus \mathbf{S}_i(\mathbf{b}_i)$ must be descendants of $\mathbf{D} \cup \pa_{\cal G}(\mathbf{D})$. To identify $p(\mathbf{D} \mid \doo(\pa_{\cal G}(\mathbf{D})))$  from $p(\mathbf{V}_i (\mathbf{b}_i))$, this involves fixing on all parents of the intrinsic set, plus all descendants of the set (i.e. marginalization of descendants). However, $\mathbf{D}$ will also be intrinsic in $p(\mathbf{S}_i (\mathbf{b}_i))$, since the interventional distribution will differ from $p(\mathbf{V}_i (\mathbf{b}_i))$ only by descendants (which have been marginalized). The ancestral property of the interventional distributions guarantees that the parents of this intrinsic set are present, in fixed or fixable form. Thus if $\mathbf{D} \cup \pa_{\cal G}(\mathbf{D})$ is present in $p(\mathbf{V}_i (\mathbf{b}_i))$, then it is also present in $p(\mathbf{S}_i (\mathbf{b}_i))$.
    Therefore, only cases 1 and 2 can exist. The implication is that the appropriate witness to non-identification is a thicket. We demonstrate this by outlining a constructive proof. 

    Assume that the aID algorithm fails on some district $\mathbf{D}$, with some set of interventional marginals $\mathbb{Z}$, and some arbitrary completion of those marginals $\bar{\mathbb Z}$. One possibility is that $\mathbf{D} \cup \pa_{\cal G} (\mathbf{D})$ falls into case 1 for each $p(\mathbf{S}_i (\mathbf{b}_i))$ and $p(\mathbf{V}_i (\mathbf{b}_i))$ respectively. In this case, the thicket construction is exactly as outlined in \citet{leeGeneralIdentifiabilityArbitrary2019}. This thicket construction will show that $p^1(\mathbf{V}_i (\mathbf{b}_i)) = p^2 (\mathbf{V}_i (\mathbf{b}_i))$ for all $i$, but the causal effects disagree - $p^1(\mathbf{D}(\pa_{\cal G}(\mathbf{D})) ) \neq p^2 (\mathbf{D}(\pa_{\cal G}(\mathbf{D})))$. Marginalizing $\mathbf{V}_i \setminus \mathbf{S}_i$ over each $p(\mathbf{V}_i (\mathbf{b}_i))$ in both ${\cal M}_1, {\cal M}_2$ gives $p^1(\mathbf{S}_i (\mathbf{b}_i)) = p^2(\mathbf{S}_i (\mathbf{b}_i))$ for all $i$, by definition. This proves non-identifiability because we have supplied two sets of distributions which agree observationally, but disagree on the desired causal effect.

    On another extreme, we may find that $\mathbf{D} \cup \pa_{\cal G} (\mathbf{D})$ falls into case 2 for some $p(\mathbf{S}_i (\mathbf{b}_i))$ and $p(\mathbf{V}_i (\mathbf{b}_i))$. In such a case, non-identification is straightforward. The thicket construction applies as we can construct $p^1(\mathbf{S}_i(\mathbf{b}_i)) = p^2(\mathbf{S}_i (\mathbf{b}_i))$ by marginalizing $\mathbf{V}_i \setminus \mathbf{S}_i$ for each $i$, but the causal effects disagree - $p^1(\mathbf{D}(\pa_{\cal G}(\mathbf{D})) ) \neq p^2 (\mathbf{D}(\pa_{\cal G}(\mathbf{D})))$.

    In practice, cases 1 and 2 may be present - but as the procedure for handling the cases is the same, the above argument extends to handling such $\mathbf{D}$.

    The proof of Lemma 3 in \citet{leeGeneralIdentifiabilityArbitrary2019} provides an explicit construction of this thicket, once we have extended $\mathbb{Z}$ to $\bar{\mathbb{Z}}$. When gID fails, it will have as arguments the causal effect $p(\mathbf{D} \mid \doo (\mathbf{V} \setminus \mathbf{D})) = p(\mathbf{D} \mid \doo(\pa_{\cal G}(\mathbf{D}))) = p(\mathbf{D}(\pa_{\cal G}(\mathbf{D})))$, graph ${\cal G}_{\mathbf{Y}^*}$, and the set of extended distributions $\mathbb{Z}$.

\end{prf}

\begin{thma}{\ref{lem:aid_complete}}
    aID is complete.

\end{thma}
\begin{prf}
    aID receives as inputs the interventional distributions $\mathbb{Z}$ and the desired causal effect $p(\mathbf{Y}(\mathbf{a}))$ in some graph $\mathcal{G}$. The algorithm will fail if for some intrinsic set $\mathbf{D}$ with parents $\pa_{\mathcal{G}}(\mathbf{D})$, the kernel $q(\mathbf{D} \mid \mathbf{V} \setminus \mathbf{D })$ cannot be obtained via a valid fixing sequence from any distribution $p(\mathbf{S}_i (\mathbf{b}_i)) \in \mathbb{Z}$, for  $\mathbf{D} \in \mathcal{D}(\mathcal{G}_{\mathbf{Y}^*})$

    We can construct the completion of the interventions, denoted $\bar{\mathbb{Z}}$. 

    If the desired intrinsic set $\mathbf{D}$ with parents $ \pa_{\cal G}(\mathbf{D}))$ fails to be found using gID, then by Lemma \ref{lem:gid_fails} we can create a thicket and models $\mathcal{M}_1, \mathcal{M}_2$ demonstrating that $p(\mathbf{Y}(\mathbf{a}))$ could not be identified.

    If the desired intrinsic set $\mathbf{D}$ with parents $\pa_{\cal G}(\mathbf{D}))$ is identified using gID, then by Lemma \ref{lem:aid} we can show that this can only happen in cases where a thicket construction applies, and that $p(\mathbf{Y}(\mathbf{a}))$ could not be identified.

    Hence, if aID fails, the causal effect is not identified. 

\end{prf}

\section{Equivalence of gID and one-line gID}

Section \ref{sec:sound} provides proof that the one-line gID algorithm is sound. It remains to show that whenever the one-line gID algorithm fails, there exists a thicket preventing identification.

Line 6 of Algorithm 1 in \cite{leeGeneralIdentifiabilityArbitrary2019} is equivalent to the expression for one-line gID (Lemma 1). Both algorithms have the following steps: there is a summation over variables which are ancestors of $\mathbf{Y}$ - guaranteed by Line 3 in Algorithm 1, and the definition of $\mathbf{Y}^*$ in Lemma 1; a product over districts in the graph $\mathcal{G}(\mathbf{a})$; and an operation which attempts to obtain $p(\mathbf{D} \mid \doo(\pa(\mathbf{D})))$ in one of the interventional distributions available. This final operation relies on the ID algorithm applied to each interventional distribution - either in its original recursive form \cite{shpitserIdentificationJointInterventional2006} or as the one-line version using the nested Markov factorization \cite{richardsonNestedMarkovProperties2017}. 

\section{Examples}
\subsection{Nested Markov factorization}
We demonstrate the nested Markov factorisation of some distribution $p(A, B, C, D)$ with corresponding graph

\begin{figure}
    \centering
        \begin{tikzpicture}[
            > = stealth, 
            auto,
            semithick 
            ]

            \tikzstyle{state}=[
            draw = none,
            fill = white,
            minimum size = 1mm
            ]

            \node[state]  (A) {$A$};
            \node[state] (B) [right of=A] {$B$};
            \node[state] (C) [right of=B] {$C$};
            \node[state] (D) [right of=C] {$D$};

            \path[->, blue] (A) edge node {} (B);
            \path[->, blue] (B) edge node {} (C);
            \path[->, blue] (C) edge node {} (D);
            \path[->, blue, bend right=30] (A) edge node {} (C);
            \path[<->, red, bend right=30] (B) edge node {} (D);

        \end{tikzpicture}
    \label{fig:nested}
    \caption{Nested Markov Factorization Example}
\end{figure}
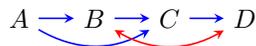

We first consider all reachable sets, of which there are $13$.  Out of these sets, $5$ are intrinsic (these are listed first), and correspond to intrinsic Markov kernels making up the nested Markov factorization of all reachable sets.  Every reachable set corresponds to a distribution equal to a particular product of intrinsic Markov kernels.
\begin{itemize}
    \item $\{A\}$, $q_A(A) \equiv p(A)$.  This is an intrinsic set.
    \item $\{B\}$, $q_B(B | A) \equiv p(B | A)$.  This is an intrinsic set.
    \item $\{C\}$, $q_C(C | B,A) \equiv p(C | B,A)$.  This is an intrinsic set.
    \item $\{D\}$, $q_D(D | C) \equiv \sum_{B} p(D | C,B,A) p(B | A)$.  This is an intrinsic set.
    \item $\{B, D\}$, $q_{B,D}(B,D | C,A) \equiv p(D | C,B,A) p(B | A)$.  This is an intrinsic set.
    \item $\{A, B\}$, $q_A(A) \cdot q_B(B | A)$.  This is a reachable (but not intrinsic) set.
    \item $\{A, C\}$, $q_A(A) \cdot q_C(C | B,A)$.  This is a reachable (but not intrinsic) set.
    \item $\{B, C\}$, $q_B(B | A) \cdot q_C(C | B,A)$.  This is a reachable (but not intrinsic) set.
    \item $\{A, D\}$, $q_D(D | C) \cdot q_A(A)$.  This is a reachable (but not intrinsic) set.
    \item $\{A, B, C\}$, $q_A(A) \cdot q_B(B | A) \cdot q_C(C | B,A)$.  This is a reachable (but not intrinsic) set.
    \item $\{A, B, D\}$, $q_A(A) \cdot q_{B,D}(B,D | C,A)$.  This is a reachable (but not intrinsic) set.
    \item $\{B, C, D\}$, $q_C(C | B,A) \cdot q_{B,D}(B,D | C,A)$.  This is a reachable (but not intrinsic) set.
    \item $\{A, B, C, D\}$, $q_A(A) \cdot q_C(C | B,A) \cdot q_{B,D}(B,D | C,A)$.  This is a reachable (but not intrinsic) set.
\end{itemize}

Note that the kernel $q_D(D | C)$ is not a function of $A$ under the model, although it may appear that it might be at first glance, based on the form of its functional in terms of $p(A,B,C,D)$, namely $\sum_B p(D \mid A, B, C)p(B \mid A)$.  This is a \emph{generalized independence constraint} or a \emph{Verma constraint} \cite{verma90equiv}.

\subsection{gID example} \label{ex:gid}
To illustrate gID reformulated using the nested Markov factorization (as given in Lemma \ref{lem:gid}), we consider the problem of identifying $p(\mathbf{Y}(\mathbf{a})) =p(Y(x_1, x_2))$ in Fig. \ref{fig:latentu1}, with access to $\mathbb{Z} = \{p(\mathbf{V}_1(\mathbf{b}_1)) = p(\{\mathbf{V}_{X_1}(x_1)), p(\mathbf{V}_2 (\mathbf{b}_2)) = p(\{\mathbf{V}_{X_2}(x_2))\}$, represented by Figs. \ref{fig:latentu2} and \ref{fig:latentu3} respectively.

The districts of ${\cal G}(\mathbf{Y}^*)$ are ${\cal D}({\cal G}(\mathbf{Y}^*) = \{ \{Y, U\}, \{W\}\}$. The required intrinsic Markov kernels will be of the form $q_{\mathbf{V}}(\mathbf{D} \mid \pa_{\cal G}(\mathbf{D}))$.
These are $q_{\mathbf{V}}(Y, U \mid W, X_2), q_{\mathbf{V}}(W \mid X_1)$.

We consider $\mathbf{D}_1 = \{Y, U\}$. This set is reachable in ${\cal G}(\mathbf{V}_{x_2}(x_2))$ by fixing $W$ and then $X_1$, so $i_{\mathbf{D}_1} = 2$. Then, the intrinsic Markov kernel can be recovered as

\begin{align*}
    &\phi_{\mathbf{V}_2 \setminus \mathbf{D}_1 }(p(\mathbf{V}_2 (\mathbf{b}_2)); {\cal G}(\mathbf{V}_2 (\mathbf{b}_2))) \vert_{\mathbf{A} = \mathbf{a}}  \\
    &= \phi_{\{X_1, W \}}(p(\mathbf{V}_{X_2}(x_2)); {\cal G}(\mathbf{V}_{X_2} (x_2))) \vert_{X_1 = x_1, X_2 = x_2}\\
    &= \phi_{X_1} \Big(\frac{p(\mathbf{V}_{X_2}(x_2))}{p(W \mid X_1, U(x_2))}; {\cal G}((\mathbf{V}_{X_2} \setminus \{W\}) (x_2)) \Big)\vert_{X_1 = x_1, X_2 = x_2}\\
    &= \phi_{X_1} \Big(p(Y(x_2)\mid X_1, U(x_2), W) p(X_1, U(x_2)); {\cal G}((\mathbf{V}_{X_2} \setminus \{W\}) (x_2)) \Big)\vert_{X_1 = x_1, X_2 = x_2}\\
    &= \phi_{X_1} \Big(q_{\mathbf{V}}(Y(x_2), X_1, U(x_2)\mid W) ; {\cal G}((\mathbf{V}_{X_2} \setminus \{W\}) (x_2)) \Big)\vert_{X_1 = x_1, X_2 = x_2}\\
    &= \frac{q_{\mathbf{V}}(Y(x_2), X_1, U(x_2)\mid W) }{q_{\mathbf{V}}(X_1 \mid  W)}\vert_{X_1 = x_1, X_2 = x_2}\\
    &= \frac{p(Y(x_2)\mid X_1, U(x_2), W) p(X_1, U(x_2))}{p(X_1)}\vert_{X_1 = x_1, X_2 = x_2} \\
    &= p(Y(x_2)\mid X_1=x_1, U(x_2), W) p(U(x_2)\mid X_1 = x_1)
\end{align*}

By similar logic, $\mathbf{D}_2 = \{W\}$ is reachable in ${\cal G}(\mathbf{V}_{X_2}(x_2))$ by fixing $Y(x_1), U, X_2$ in that order, so $i_{\mathbf{D}_2} = 1$. The reader can verify that the intrinsic Markov kernel is 
\begin{align*}
    &\phi_{\mathbf{V}_1 \setminus \mathbf{D}_2 } (p(\mathbf{V}_1 (\mathbf{b}_1)); {\cal G}(\mathbf{V}_1(\mathbf{b}_1)))\vert_{\mathbf{A}= \mathbf{a}}\\
    &= \phi_{\{Y(x_1), U, X_2\}}(p(\mathbf{V}_{X_1}(x_1));{\cal G}(\mathbf{V}_{X_1} (x_1)))\vert_{\mathbf{A} = \mathbf{a}} \\
    &=q_{\mathbf{V}}(W \mid X_1, U(x_2), W) \\
    &= p(W(x_1) )
\end{align*}

Hence, $p(Y(x_1, x_2))$ is 
\begin{align*}
&\sum_{\mathbf{Y}^* \setminus \mathbf{Y}} \prod_{\mathbf{D} \in {\cal D}({\cal G}({\mathbf{Y}^*}(\mathbf{a})))}
    \phi_{\mathbf{V}_{i_{\mathbf{D}}} \setminus \mathbf{D}}(p(\mathbf{V}_{i_{\mathbf{D}}}(\mathbf{b}_{i_{\mathbf{D}}})); {\cal G}(\mathbf{V}_{i_{\mathbf{D}}}(\mathbf{b}_{i_{\mathbf{D}}}))) \vert_{\mathbf{A} = \mathbf{a}}\\
&= \sum_{W, U} p(Y(x_2)\mid X_1=x_1, U(x_2), W) p(U(x_2)\mid X_1 = x_1)p(W(x_1) ),\\
&= \sum_{W, U} p(Y(x_2)\mid U(x_2), W) p(U(x_2))p(W(x_1) ).
\end{align*}
where the last equality holds by the Markov properties, since $Y \ci X_1 \mid U, W$, and $U \mid X_1$ in ${\cal G}(\{\mathbf{V} \setminus \{X_2\}\})$, represented by Fig. \ref{fig:latentu2}.

\subsection{aID example} \label{ex:aid}

To illustrate the aID algorithm presented in Lemma \ref{lem:aid}, consider again the problem of identifying $p(Y(x_1, x_2))$ in Fig.~\ref{fig:latentu1}, where 
\begin{align*}
    \mathbb{Z}= &\{p(\mathbf{S}_1(\mathbf{b}_1)) = p(W(x_1)), \\
                &\quad p(\mathbf{S}_2(\mathbf{b}_2)) = p(\{W, U, Y, X_1\}(x_2)) \},
\end{align*}
 corresponding to Figs.~\ref{fig:latentu4} and \ref{fig:latentu3} respectively. Marginal distributions are not considered valid inputs to gID, causing it to trivially fail. However, the intrinsic Markov kernels might still be identified.

The districts of ${\cal G}(\mathbf{Y}^*)$ are ${\cal D}({\cal G}(\mathbf{Y}^*) = \{ \{Y, U\}, \{W\}\}$. The required intrinsic Markov kernels will be of the form $q_{\mathbf{V}}(\mathbf{D} \mid \pa_{\cal G}(\mathbf{D}))$.
These are $q_{\mathbf{V}}(Y, U \mid W, X_2), q_{\mathbf{V}}(W \mid X_1)$.

Considering $\mathbf{D}_1 = \{Y, U\}$, we notice that this set is reachable in ${\cal G}(\mathbf{S}_2 (\mathbf{b}_2))$ by fixing $W(x_2)$, and then $X_1$. The corresponding intrinsic Markov kernel is 
\[q_{\mathbf{V} \setminus \{Y(x_2), U(x_2)\}}(Y(x_2), U(x_2) \mid W(x_2)),\]

which is equal to
\begin{align*}
    &\phi_{\mathbf{S}_2 \setminus \mathbf{D}_1} (p(\mathbf{S}_2 (\mathbf{b}_2)); {\cal G}(\mathbf{S}_2 (\mathbf{b}_2))) \vert_{\mathbf{A} = \mathbf{a}}\\
    &=\phi_{X_1, W} (p(\mathbf{S}_2 (\mathbf{b}_2)); {\cal G}(\mathbf{S}_2 (\mathbf{b}_2))) \vert_{\mathbf{A} = \mathbf{a}}\\
    &= \phi_{X_1}(p(Y(x_2) \mid W, U(x_2), X_1) p(U(x_2), X_1); \phi_{W}{\cal G}(\mathbf{S}_2 (\mathbf{b}_2))), \\
    &= \frac{p(Y(x_2) \mid W, U(x_2), X_1) p(X_1 ,U(x_2) ) }{p(X_1)} \\
    &= p(Y(x_2) \mid W, U(x_2), X_1) p( U(x_2)  \mid X_1)
\end{align*}
$\mathbf{D}_2 = \{W\}$ is reachable in ${\cal G}(\mathbf{S}_1 (\mathbf{b}_1))$, without any fixing operations. In this case,
\begin{align*}
    q_{\mathbf{V} \setminus \{W\}}(W(x_1)) &= \phi_{\mathbf{S}_1 \setminus \mathbf{D}_2 } (p(W(x_1)) ; {\cal G}(\mathbf{S}_1 (\mathbf{b}_1))) \vert_{\mathbf{A} = \mathbf{a}} \\
&= p(W(x_1)) \\
\end{align*}
since $\mathbf{S}_1 \setminus \mathbf{D}_2 = \emptyset$.

Therefore, $p(Y(x_1, x_2))$ is 

\[\sum_{W, U} p(Y(x_2) \mid W, U(x_2), X_1) p(U(x_2)\mid X_1) \times p(W(x_1)) \]

\subsection{mID example}\label{ex:mid}
Once again, we return to the problem of identifying $p(Y(x_1, x_2))$ in Fig. \ref{fig:latentu1}, where our interventional distributions are 
\begin{align*}
    \mathbb{Z} =& \{p(\mathbf{S}_1(\mathbf{b}_1)) = p(W(x_1)) \\
                & \quad p(\mathbf{S}_2(\mathbf{b}_2)) = p(\{Y, W\}(x_2))\}
\end{align*}

given by Figs. \ref{fig:latentu4} and \ref{fig:latentu6}. This differs from the aID example because the marginal distribution represented by ${\cal G}(\mathbf{S}_2(\mathbf{b}_2))$ is not ancestral with respect to ${\cal G}(\mathbf{V}(x_1, x_2))$.

The algorithm is functionally identical to aID, with the exception that it is applied over each $\mathbf{Y}' \subseteq {\cal P}(\mathbf{Y}^* \setminus \mathbf{Y}) \cup \mathbf{Y}$

We first verify that the algorithm fails when $\mathbf{Y}' \equiv \mathbf{Y}^*$. In this case, the districts of ${\cal G}(\mathbf{Y}')$ are ${\cal D}({\cal G}(\mathbf{Y}')) = \{ \{Y, U\}, \{W\}\}$. We immediately see that since $U$ is not available in $\mathbb{Z}$, the algorithm will fail.

Then, we consider $\mathbf{Y}' \equiv \mathbf{Y}^* \setminus \{U\}$. The distrcts of ${ \cal G } (\mathbf{Y}')$ are ${\cal D}({\cal G}(\mathbf{Y}')) = \{ \{Y\}, \{W\}\}$. Since all variables of ${\cal D}({\cal G}(\mathbf{Y}')) $ are present in $\mathbb{Z}$, it is possible that it is identified. 

For $\mathbf{D}_1 = \{Y\}$, this is reachable in ${\cal G}(\mathbf{S}_2(\mathbf{b}_2))$, and the corresponding kernel is given by
\begin{align*}
    &\phi_{\mathbf{S}_2 \setminus \mathbf{D}_1 } (p(\{Y, W\}(x_2)); {\cal G}(\{Y, W\}(x_2)) )\vert_{X_1 = x_1, X_2 = x_2} \\
    &=\phi_{\{W\}} (p(\{Y, W\}(x_2)); {\cal G}(\{Y, W\}(x_2)) ) \vert_{X_1 = x_1, X_2 = x_2}\\
    &= \frac{p(Y(x_2), W)}{p(W)} \vert_{X_1 = x_1, X_2 = x_2}\\
    &= p(Y(x_2) \mid W))
\end{align*}

The procedure for obtaining the kernel associated with $\mathbf{D}_2 = \{W\}$ is exactly as described in the aID example. The corresponding kernel is
\[p(W(x_1))\]

Therefore, $p(Y(x_1, x_2))$ is identified as 
\[\sum_{W} p(Y(x_2)\mid W) p(W(x_1)).\]

\subsection{eID example}\label{ex:eid}

\begin{figure*}
    \centering
    \begin{subfigure}[b]{0.3\textwidth}
        \begin{tikzpicture}[
            > = stealth, 
            auto,
            semithick 
            ]

            \tikzstyle{state}=[
            draw = none,
            fill = white,
            minimum size = 1mm
            ]
            \node[state] at (0, 0) (X1) {$X_1$};
            \node[state] at (3, 0) (X2)  {$X_2$};
            \node[state] at (0, -1) (W1)  {$W_1$};
            \node[state] at (0, -2) (R1)  {$R_1$};
            \node[state] at (1, -1) (J)  {$J$};
            \node[state] at (2, -1) (C)  {$C$};
            \node[state] at (3, -1) (W2)  {$W_2$};
            \node[state] at (3, -2) (R2)  {$R_2$};

            \path[->, blue] (X1) edge node {} (W1);
            \path[->, blue] (X2) edge node {} (W2);
            \path[->, blue] (W2) edge node {} (R2);
            \path[->, blue] (W1) edge node {} (R1);
            \path[->, blue] (J) edge node {} (C);
            \path[->, blue] (J) edge node {} (W1);


            \path[<->, red, bend right=30] (X1) edge node {} (W1);
            \path[<->, red, bend left=30] (X2) edge node {} (W2);
            \path[<->, red, bend right=30] (W1) edge node {} (R1);
            \path[<->, red] (W2) edge node {} (C);
            \path[<->, red] (R1) edge node {} (R2);

        \end{tikzpicture}
        \caption{$\mathcal{G}(\mathbf{V})$}
        \label{fig:eid1}
    \end{subfigure}
    \begin{subfigure}[b]{0.3\textwidth}
        \begin{tikzpicture}[
            > = stealth, 
            auto,
            semithick 
            ]

            \tikzstyle{state}=[
            draw = none,
            fill = white,
            minimum size = 1mm
            ]
            \tikzstyle{fixed}=[
            draw=black,
            rectangle, 
            thick,
            fill = white,
            minimum size = 1mm
            ]

            \node[fixed] at (0, 0) (X1) {$X_1$};
            \node[state] at (3, 0) (X2)  {$X_2$};
            \node[state] at (0, -1) (W1)  {$W_1$};
            \node[state] at (0, -2) (R1)  {$R_1$};
            \node[state] at (1, -1) (J)  {$J$};
            \node[state] at (2, -1) (C)  {$C$};
            \node[state] at (3, -1) (W2)  {$W_2$};
            \node[state] at (3, -2) (R2)  {$R_2$};

            \path[->, blue] (X1) edge node {} (W1);
            \path[->, blue] (X2) edge node {} (W2);
            \path[->, blue] (W2) edge node {} (R2);
            \path[->, blue] (W1) edge node {} (R1);
            \path[->, blue] (J) edge node {} (C);
            \path[->, blue] (J) edge node {} (W1);
            \path[<->, red, bend left=30] (X2) edge node {} (W2);
            \path[<->, red, bend right=30] (W1) edge node {} (R1);
            \path[<->, red] (W2) edge node {} (C);
            \path[<->, red] (R1) edge node {} (R2);


        \end{tikzpicture}
        \caption{$\mathcal{G}((\mathbf{V} \setminus \{X_1\})(x_2))$ }
        \label{fig:eid2}

    \end{subfigure}
    \begin{subfigure}[b]{0.3\textwidth}
        \begin{tikzpicture}[
            > = stealth, 
            auto,
            semithick 
            ]

            \tikzstyle{state}=[
            draw = none,
            fill = white,
            minimum size = 1mm
            ]

            \tikzstyle{fixed}=[
            draw=black,
            rectangle, 
            thick,
            fill = white,
            minimum size = 2mm
            ]

            \node[fixed] at (1, 1) (X2)  {$X_2$};
            \node[state] at (-1, 0) (J)  {$J$};
            \node[state] at (-1, -1) (R1)  {$R_1$};
            \node[state] at (0, 0) (C)  {$C$};
            \node[state] at (1, 0) (W2)  {$W_2$};
            \node[state] at (1, -1) (R2)  {$R_2$};

            \path[->, blue] (X2) edge node {} (W2);
            \path[->, blue] (W2) edge node {} (R2);
            \path[->, blue] (J) edge node {} (C);
            \path[<->, red] (W2) edge node {} (C);
            \path[<->, red] (R1) edge node {} (R2);

        \end{tikzpicture}
        \caption{$\mathcal{G}(\{R_1, R_2, C, W_2\}(x_2))$}
        \label{fig:eid3}
    \end{subfigure}
    \label{fig:eid}
    \caption{Example for eID}
\end{figure*}
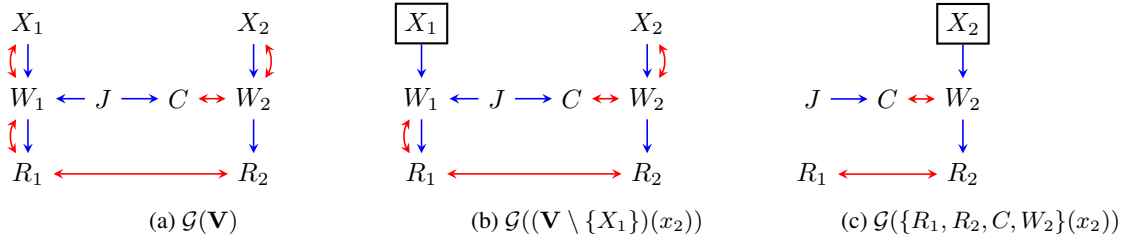

To demonstrate eID using Lemma \ref{lem:eid_statement}, we consider identifying $p(\{R_1, R_2\}(x_1, x_2))$ in $\mathcal{G}$ represented by Figure \ref{fig:eid1} from conditional marginal interventional distributions represented by Figs. \ref{fig:eid2} and \ref{fig:eid3}. In particular,
\begin{align*}
    \mathbb{Z}' &= \{p(\{J, R_1, W_2, X_2\}(x_1) \mid C(x_1)=c, \{W_1, R_1\}(x_1)), \\
                & \qquad p(R_1(x_1) \mid W_1(x_1)),\\
                &\qquad p(W_1(x_1) \mid C(x_1) = c),\\
                &\qquad p(\{J, R_1, R_2, W_2, C\}(x_2))\}
\end{align*}

where $\mathbb{Z}'$ has not been chain rule closed. Unless otherwise specified, we assume that we have conditional distributions available at all levels of conditioning. 

Recall that to make a set of distributions chain rule closed, it may be necessary to apply the rules of po-calculus \cite{malinskyPotentialOutcomesCalculus2019}. In particular, we can apply rule 1 to $p(R_1 (x_1) \mid W_1(x_1))$. Since $R_1 (x_1) \ci C(x_1) \mid W_1(x_1)$ in ${\cal G}(\{\mathbf{V} \setminus \{X_1\})(x_1))$, rule 1 states that 
\begin{align*}
    p(R_1(x_1) \mid W_1(x_1) ) &= p(R_1(x_1) \mid W_1(x_1), C(x_1))\\
                               &= p(R_1 (x_1) \mid W_1(x_1), C(x_1) = c))
\end{align*}

Then, applying chain rule we notice that the first three distributions may be combined in the following manner:
\begin{align*}
    p(\{J, R_1, R_2, W_1, W_2, X_2\}(x_1) \mid C(x_1) = c) &= p(\{J, R_1, W_2, X_2\}(x_1) \mid C(x_1)=c, \{W_1, R_1\}(x_1)) \\
                                                        &\qquad \times p(R_1 (x_1) \mid W_1(x_1), C(x_1) = c))\\
                                                        &\qquad \times p(W_1(x_1) \mid C(x_1) = c)
\end{align*}

After one chain rule closure step, we obtain
\begin{align*}
    \mathbb{Z} = &\{p(\mathbf{S}_1(\mathbf{b}_1) \mid \mathbf{C}_1(\mathbf{b}_1) = \mathbf{c}_1 ) = p(\{J, R_1, R_2, W_1, W_2, X_2\}(x_1) \mid C(x_1) = c), \\
        &p(\mathbf{S}_2(\mathbf{b}_2) \mid \mathbf{C}_2(\mathbf{b}_2) = \mathbf{c}_2 )=p(\{J, R_1, R_2, W_2, C\}(x_2) ),\\
        &p(\{J, R_1, W_2, X_2\}(x_1) \mid C(x_1)=c, \{W_1, R_1\}(x_1)), \\
        &p(R_1(x_1) \mid W_1(x_1)),\\
        &p(W_1(x_1) \mid C(x_1) = c),\\
        &\ldots
    \}
\end{align*}
where we recognise that there may be other distributions attainable by further chain rule closures, but that only the first two interventional distributions are sufficient for this problem.

Ordinarily, the algorithm iterates over $\mathbf{Y}' \in {\cal P}(\mathbf{Y}^* \setminus \mathbf{Y}) \cup \mathbf{Y}$. For $\mathbf{Y}' \equiv \mathbf{Y}^* = \{J, R_1, R_2, W_2, W_1\}$, the relevant districts are ${\cal D}({\cal G}(\mathbf{Y}^*))=\{R_1, R_2, W_1\} ,\{W\}, \{J\}\}$. In this example we will find that it is sufficient to consider $\mathbf{Y}^*$ only, but this is not true in general as per the mID example.

We now consider the problem of identifying $p(\mathbf{D}_1 \mid \doo(\pa_{\cal G}(\mathbf{D}_)))$ using Lemma \ref{lem:joint-from-conditional}. For this indented block, assume $\mathbf{Y}^*$ to be scoped with reference to Lemma \ref{lem:joint-from-conditional}, and not Lemma \ref{lem:eid_statement}.
\begin{quote}
For $\mathbf{D}_1 = \{R_1, R_2, W_1\}$, we find that the corresponding Markov kernel $p(\mathbf{D}_1 \mid \doo(\pa_{\cal G} (\mathbf{D}_1)))$ is identified from $p(\mathbf{S}_1(\mathbf{b}_1) \setminus \mathbf{C}_1 \mid \mathbf{C}_1(\mathbf{b}_1) =\mathbf{c}_1)) = p(\{J, R_1, R_2, W_1, W_2, X_2\}(x_1) \mid C(x_1) = c)$, using Lemma \ref{lem:joint-from-conditional}. We find that all consistency conditions in the Lemma are satisfied. Further, $\mathbf{Y}^* = \{\{R_1, R_2, W_1\}\}$. For this district $\mathbf{D} = \{R_1, R_2, W_1\}$, we consider $\mathbf{Z}_{\mathbf{D}} = \{W_2, X_2, R_2, W_1, R_1\}$, and $\bar{\mathbf{D}} = \mathbf{D}$, and show that these are suitable candidates (in general, one would need to search over all such candidates $\mathbf{Z}_{\mathbf{D}}$ to verify if they were suitable). First, $\mathbf{Z}_{D} \in \mathbf{S} \setminus \mathbf{C} = \{R_1, R_2, W_1, W_2, X_2\}$. Second, $\mathbf{D} \subseteq \mathbf{Z}_{\mathbf{D}}$. Third, $\mathbf{Z}_{\mathbf{D}}$ is s-fixable in $\mathcal{G}(\{\mathbf{V} \setminus \{X_1\}\}(x_1))$ with a sequence that fixes $\bar{\mathbf{D}}$ last -- we may fix $\mathbf{Z}_{\mathbf{D}} \setminus \mathbf{D} = \{W_2, X_2\}$, and then members of $\bar{\mathbf{D}} = \{R_2, R_1, W_1\}$ in the order presented. Fourth, one can verify that $\bar{\mathbf{D}}$ is a district in $\phi^{\mathbf{C}}_{\sigma_{\mathbf{Z}_{\mathbf{D}}} \setminus \bar{\mathbf{D}}} (\mathcal{G}(\{\mathbf{V} \setminus \{X_1\}\}(x_1)))$ -- fixing $\mathbf{Z}_{\mathbf{D}} = \{W_2, X_2\}$ does not disrupt the district $\mathbf{D}$. Fifth, $\mathbf{D}$ is reachable in $\phi^{\mathbf{C}}_{\sigma_{\mathbf{Z}_{\mathbf{D}}} \setminus \bar{\mathbf{D}}} (\mathcal{G}(\{\mathbf{V} \setminus \{X_1\}\}(x_1)))$ since $\mathbf{D} = \bar{\mathbf{D}}$. Then, applying the rest of the Lemma, we note that for our single $\mathbf{D}$:
\begin{align*}
    q_{\bf D}({\bf D} | \pa^s_{{\cal G}({\bf S}({\bf b}))}({\bf D})) &= \phi_{\bar{\bf D} \setminus {\bf D}}(q_{\bar{\bf D}}(\bar{\bf D} | \pa^s_{{\cal G}}(\bar{\bf D}));
    \phi_{{\bf S} \setminus \bar{\bf D}}({\cal G}({\bf S}({\bf b}))))
\end{align*}
does not do anything since $\mathbf{D} = \bar{\mathbf{D}}$.

Then, 
\begin{align*}
&q_{{\bf S} \setminus ({\bf Z}_{\bf D} \!\setminus\! \bar{\bf D})}({\bf S} \!\setminus\! ({\bf Z}_{\bf D} \!\setminus\! \bar{\bf D}) | {\bf Z}_{\bf D} \!\setminus\! \bar{\bf D})\\
&\equiv \phi^{\bf C}_{{\bf Z}_{\bf D} \!\setminus\! \bar{\bf D}}(p({\bf S}({\bf b} \mid {\bf C}({\bf b}) = {\bf c}); {\cal G}({\bf S}({\bf b})))) \\
&=\phi^{\mathbf{C}}_{X_2, W_2} (p((\{\mathbf{V} \setminus \{X_1, C\}\} (x_1) \mid C(x_1) = c);\mathcal{G}(\{\mathbf{V} \setminus \{X_1\}\}(x_1))) \\
&=\phi^{\mathbf{C}}_{X_2} (\frac{p(\{R_1, R_2, W_1, W_2, J, X_2\}(x_1) \mid C(x_1) = c)}{p(W_2 \mid C(x_1) = c, J, R_1, W_1, X_2)};\phi^{C}_{W_2}\mathcal{G}(\{\mathbf{V} \setminus \{X_1\}\}(x_1))) \\
&=\phi^{\mathbf{C}}_{X_2} (p(\{R_2\}(x_1) \mid C(x_1) = c, \{R_1, W_1, W_2, X_2, J \}(x_1)) p(\{R_1, W_1 , X_2, J\}(x_1) \mid C(x_1) = c);\phi^{C}_{W_2}\mathcal{G}(\{\mathbf{V} \setminus \{X_1\}\}(x_1)))\\
\end{align*}
Hence,
\begin{align*}
&q_{R_1, R_2, W_1, J} (R_1, R_2, W_1, J \mid W_2, X_2)\\
&= p(\{R_2 \}(x_1) \mid C(x_1) = c, \{R_1, W_1, W_2, X_2, J\}(x_1)) p(\{R_1, W_1,  J\}(x_1) \mid C(x_1) = c)
\end{align*}
Finally, we note that for the topological order $W_1 \prec R_1 \prec R_2$, 
\begin{align*}
    q_{\bar{\bf D}}(\bar{\bf D} | \pa^s_{{\cal G}}(\bar{\bf D})) &\equiv \prod_{D \in \bar{\bf D}} q_{{\bf S} \setminus ({\bf Z}_{\bf D} \setminus \bar{\bf D})}(D | \mb^*(D), {\bf Z}_{\bf D} \setminus \bar{\bf D})\\
&= q_{R_1, R_2, W_1, J} (W_1\mid  J,  W_2, X_2) q_{R_1, R_2, W_1, J} (R_1\mid W_1, J,  W_2, X_2) q_{R_1, R_2, W_1, J} (R_2\mid R_1, W_1, J,  W_2, X_2)  \\
\end{align*}
\textcolor{red}{
    Expanding each of these factors by appealing to the definition of conditioning in a kernel, we obtain
\begin{align*}
    q_{R_1, R_2, W_1, J}(W_1 \mid J, W_2, X_2)    &= \sum_{R_1, R_2}\frac{q_{R_1, R_2, W_1, J} (R_1, R_2, W_1, J \mid W_2, X_2)}{\sum_{R_1, R_2, W_1} q_{R_1, R_2, W_1, J} (R_1, R_2, W_1, J \mid W_2, X_2)}\\
                                                  &= \sum_{R_1, R_2}\frac{p(R_2 (x_1) \mid C(x_1) = c, \{R_1, W_1, W_2, X_2, J\}(x_1)) p(\{R_1, W_1,  J\}(x_1) \mid C(x_1) = c)}{p(J(x_1) \mid C(x_1) = c)}\\
                                                  &= p(W_1(x_1) \mid C(x_1)=c, J(x_1)) \\
    q_{R_1, R_2, W_1, J} (R_1, \mid W_1, J, W_2, X_2) &= \sum_{R_2} \frac{q_{R_1, R_2, W_1, J} (R_1, R_2, W_1, J \mid W_2, X_2)}{\sum_{R_1, R_2}q_{R_1, R_2, W_1, J} (R_1, R_2, W_1, J \mid W_2, X_2)} \\
                                                      &= \sum_{R_2} \frac{p(R_2 (x_1) \mid C(x_1) = c, \{R_1, W_1, W_2, X_2, J\}(x_1)) p(\{R_1, W_1,  J\}(x_1) \mid C(x_1) = c)}{p(\{W_1, J\}(x_1) \mid C(x_1) = c)}\\
                                                      &= p(R_1(x_1) \mid C(x_1) = c, \{W_1, J\}(x_1))\\
    q_{R_1, R_2, W_1, J} (R_2 \mid R_1, W_1, J, W_2, X_2) &=  \frac{q_{R_1, R_2, W_1, J} (R_1, R_2, W_1, J \mid W_2, X_2)}{\sum_{R_2}q_{R_1, R_2, W_1, J} (R_1, R_2, W_1, J \mid W_2, X_2)} \\
                                                          &= \frac{p(R_2 (x_1) \mid C(x_1) = c, \{R_1, W_1, W_2, X_2, J\}(x_1)) p(\{R_1, W_1,  J\}(x_1) \mid C(x_1) = c)}{p(\{R_1, W_1, J\}(x_1) \mid C(x_1) = c)}\\
                                                          &= p(R_2 (x_1) \mid C(x_1) = c, \{R_1, W_1, W_2, X_2, J\}(x_1)) 
\end{align*}
which means that 
\begin{align*}
    p(\mathbf{D}_1 \mid \doo(\pa_{\cal G} (\mathbf{D}_1))) &= p(W_1(x_1) \mid C(x_1)=c, J(x_1)) \\
                                                           &\quad \times p(R_1(x_1) \mid C(x_1) = c, \{W_1, J\}(x_1)) \\
                                                           &\quad \times p(R_2 (x_1) \mid C(x_1) = c, \{R_1, W_1, W_2, X_2, J\}(x_1)) 
\end{align*}
}
\end{quote}
As Lemma \ref{lem:joint-from-conditional} is functionally equivalent to the RC algorithm presented in \citet{bareinboimRecoveringCausalEffects2015}, the reader should be able to recover the same kernel for $\mathbf{D}_1$ either way.

For $\mathbf{D}_2 = \{W_2\}$, we find that its corresponding Markov kernel is identified from $p(\mathbf{S}_2 (\mathbf{b}_2) \setminus \mathbf{C}_2 \mid \mathbf{C}_2 (\mathbf{b}_2) = \mathbf{c}_2) = p(\{R_1, R_2, W_2, J , C\}(x_2)  )$, in a straightforward manner as there are no conditionals. Applying mID techniques, 
\begin{align*}
    &p(\mathbf{D}_2 \mid \doo(\pa_{{\cal G}(\mathbf{S}_2(\mathbf{b}_2) \cup \mathbf{C}_2 (\mathbf{b}_2))}(\mathbf{D}_2)))\\
    &= p(W_2 \mid \doo(X_2))\\
    &= \phi_{R_1, R_2, J, C} (p(\{R_1, R_2, W_2, J, C\}(x_2) ) = c); {\cal G}(\mathbf{S}_2(\mathbf{b}_2))) \\
    &= \sum_{R_2, R_2, J, C} p(\{R_1, R_2, W_2, J, C\}(x_2) )\\
    &= p(W_2(x_2))
\end{align*}

For $\mathbf{D}_3 = \{J\}$, we note that $p(J \mid \doo(\pa_{\cal G} J)) =p(J)$ is obtained from $p(\mathbf{S}_2 (\mathbf{b}_2) \setminus \mathbf{C}_2 \mid \mathbf{C}_2 (\mathbf{b}_2) = \mathbf{c}_2) = p(\{R_1, R_2, W_2, J , C\}(x_2))$ by
\[p(J)= \sum_{R_2, R_2, W_2, C} p(\{R_1, R_2, W_2, J, C\}(x_2) )\]

Applying the primary equation of Lemma \ref{lem:eid_statement}, the causal effect $p(\{R_1, R_2\} (x_1, x_2))$ is given by
\begin{align*}
    &\sum_{W_1, W_2, J}\Big(p(W_1(x_1) \mid C(x_1)=c, J(x_1)) \\
    &\quad \times p(R_1(x_1) \mid C(x_1) = c, \{W_1, J\}(x_1)) \\
    &\quad \times p(R_2 (x_1) \mid C(x_1) = c, \{R_1, W_1, W_2, X_2, J\}(x_1)) 
\Big)\Big(p(W_2(x_2)) \Big)\Big( p(J)\Big)
\end{align*}

\section{Amending the Thicket Construction}
Throughout this paper we rely on the \emph{thicket} as a structure which is a witness to the non-identification of causal effects. The thicket was originally proposed in Definition 6 of \citet{leeGeneralIdentifiabilityArbitrary2019}, along with an explicit construction relying on the thicket of two models demonstrating non-identifiability of a query. We note that whilst the construction is in principle correct, there exists a minor edge case which we address now.

The construction proposed by \citet{leeGeneralIdentifiabilityArbitrary2019} only requires that the root set $\mathbf{R}$ be non-empty, but in the construction of the two models it is also required that $\mathbf{U}'^R$ ( the UCs connected to $\mathbf{R}$ in the thicket) also be non-empty. It is only possible to have bidirected edges between root nodes if there are at least \emph{two} root nodes. Therefore, the thicket construction of two models agreeing on the observed distribution $p(\mathbf{V})$ whilst disagreeing on $p(\mathbf{R}(\mathbf{a}))$ is not defined if $\mathbf{R}$ contains only one element, since bidirected edges are not defined on a single node. Since we cannot assume the cardinality of $\mathbf{R}$ in general, we amend the construction accordingly in this section.

Let $\mathcal{G}$ be a graph where $\mathbf{R}$ is a single node $R$. We delete $R$, and in its place create $\tilde{R}_1$ inheriting all directed edges, and $\tilde{R}_2$ inheriting all bidirected edges. $\tilde{R}_1$ and $\tilde{R}_2$ are connected by a bidirected edge so that $\tilde{\mathbf{R}} = \{\tilde{R}_1, \tilde{R}_2\}$ is a valid root set. The resulting graph $\tilde{\mathcal{G}}$ is a valid thicket, since it remains one district, each hedge has variables with only one child (where hedges are drawn from the original hedges in $\mathcal{G}$), and the root set is a minimal district. 

We can apply the existing construction to this thicket as $\mathbf{U}'^R$ now contains a single bidirected edge between $\tilde{R}_1$ and $\tilde{R}_2$. We define $R_n = \tilde{R}_1 \times \tilde{R}_2$, where $\times$ denotes the cartesian product, and treat it as a single node which inherits all edges in $\tilde{\mathbf{R}}$. Define this new graph as $\mathcal{G}^\times$.

\begin{lem}
    If ${\cal M}_1$ and ${\cal M}_2$ defined on $\tilde{\mathcal{G}}$ agree on available distributions and disagree on a causal effect, then so do ${\cal M}_1$ and ${\cal M}_2$ defined on $\mathcal{G}^\times$.
\end{lem}
\begin{prf}
    Consider ${\cal M}_1$, ${\cal M}_2$. By construction they disagree on $p(\mathbf{R}(\mathbf{t}'=0) = 0)$ where $\mathbf{T}' \subseteq \mathbf{T}$, such that all hedgelets are intervened upon. Then $p^1(\mathbf{R}(\mathbf{t}' = 0) = 0) > p^2(\mathbf{R}(\mathbf{t}'=0) = 0)$. Since the cartesian product is an injective map, it must be the case that $p^1(R_n(\mathbf{t}' = 0) = 0) > p^2(R_n(\mathbf{t}'=0)=0)$. 
    Additionally, ${\cal M}_1, {\cal M}_2$ also agree on any distribution $p(\mathbf{R}(\mathbf{t}'=0) = 0)$ where $\mathbf{T}' \subseteq \mathbf{T}$, such that at least one hedgelet is \emph{not} intervened upon. Again, by the injectivity of the cartesian product, it follows that $p^1(R_n(\mathbf{t}'=0) = 0) = p^2(R_n(\mathbf{t}'=0) = 0)$.
\end{prf}

\bibliography{library}
\bibliographystyle{icml2019}

\appendix